\documentclass[runningheads]{llncs}
\usepackage[T1]{fontenc}
\usepackage{graphicx}

\usepackage{booktabs}
\usepackage{multirow}
\usepackage{multicol}
\usepackage{makecell}
\usepackage{array}
\usepackage{amssymb}
\usepackage{amsmath}
\usepackage{amsfonts}
\usepackage{xcolor}
\usepackage{nicefrac}
\begin{document}
%
\title{On Learnable Parameters of Optimal and Suboptimal Deep Learning Models}

%
%

\author{
Ziwei Zheng\inst{1}  \and
Huizhi Liang\inst{1} \and
Vaclav Snasel\inst{2} \and
Vito Latora\inst{3} \and
Panos Pardalos\inst{4} \and
Guiseppe Nicosia\inst{5}\and
Varun Ojha\inst{1}
}
\authorrunning{Z. Zheng et al.}
\institute{Newcastle University, Newcastle, UK \and
Technical University of Ostrava, Ostrava, Czech Republic \and
Queen Mary University of London, UK \and
University of Florida, FL, USA \and
University of Catania, Catania, Italy}
\maketitle              
\begin{abstract}
We scrutinize the structural and operational aspects of deep learning models, particularly focusing on the nuances of learnable parameters (weight) statistics, distribution, node interaction, and visualization. By establishing correlations between variance in weight patterns and overall network performance, we investigate the varying (optimal and suboptimal) performances of various deep-learning models. Our empirical analysis extends across widely recognized datasets such as MNIST, Fashion-MNIST, and CIFAR-10, and various deep learning models such as deep neural networks (DNNs), convolutional neural networks (CNNs), and vision transformer (ViT), enabling us to pinpoint \textit{characteristics of learnable parameters} that correlate with successful networks. Through extensive experiments on the diverse architectures of deep learning models, we shed light on the critical factors that influence the functionality and efficiency of DNNs. Our findings reveal that successful networks, irrespective of datasets or models, are invariably similar to other successful networks in their converged weights statistics and distribution, while poor-performing networks vary in their weights. In addition, our research shows that the learnable parameters of widely varied deep learning models such as DNN, CNN, and ViT exhibit similar learning characteristics. 

\end{abstract}

\section{Introduction}\label{sec:intro}
Deep learning has achieved impressive results in the fields of computer vision, natural language processing, and speech recognition. From face recognition~\cite{schroff2015facenet} to autonomous driving~\cite{geiger2012we}, to medical analysis~\cite{shen2017deep}, the deep learning models have widely been used in various significant tasks. In particular, deep learning models can outperform human experts in many application scenarios. However, the lack of understanding of deep learning models, such as deep neural networks (DNNs), convolution neural networks (CNNs), and vision transformers (ViTs), has caused widespread concern and controversy~\cite{szegedy2013intriguing}. For users, deep learning models are mysterious black boxes whose decision-making processes are difficult to explain and understand. This opacity can bring unpredictable risks in actual mission-critical situations, especially in areas sensitive to security. For example, opaque automated medical diagnosis models may produce incorrect treatment recommendations, posing a threat to patients' health~\cite{shen2017deep}.

We propose a novel methodology to perform comprehensive experiments to explore the reasons behind the varying performances of deep learning models, particularly focusing on the \textit{learnability} of weight matrices in models that yield different accuracy rates on similar and varied network architectures. Our investigation delves into the characteristics of the model's learnable parameters to understand why models with similar or differing architectures can result in such varied performances. We analyze trained networks such as the gravity of weight (i.e., average weight and their distribution), node strength differences, and visualization of their position in high dimensional space using t-SNE mapping. 

Our approach involves conducting experiments on three pattern recognition datasets MNIST, Fashion-MNIST, and CIFAR-10, using a range of network architectures such as DNN, CNN, and ViT networks. These experiments are designed to reveal how the weight distributions in neural networks (NN) relate to their learning efficiency and decision-making capabilities. The main contributions of this paper are as follows:
\begin{itemize}
    \item We present a novel methodology for a comprehensive empirical study for identifying the critical role of converged weights and node strengths in characterizing the uncertainty of deep learning success and failure.
    \item We perform a visual analysis of deep learning converged weights and node strength that help characterize the optimal and suboptimal networks.
    \item Our study investigates common factors among the learnable parameters of varied deep learning architectures that process datasets varyingly in making pattern recognition, such as DNN, CNN, and ViT.    
    \item Our findings reveal that successful networks, irrespective of tasks or models, are invariably similar to other successful networks in their converged weights, while poor-performing networks vary in their weights.
\end{itemize}

\section{Related work}\label{sec:realted_work}

The analysis of NNs has unveiled insights into their performance and optimization. Voita et al.~\cite{voita2019analyzing} investigate the redundancy in Transformer architectures' multi-head self-attention mechanism, while Frankle and Carbin~\cite{frankle2018lottery} identify `lottery ticket' initialization in weight matrices that could predict network performance success. These studies emphasize the significance of weight matrix characteristics, a theme our research echoes by analyzing how these structural elements influence learning outcomes.

Further exploration by Neyshabur et al.~\cite{neyshabur2018towards} into over-parameterization and its impact on model generalization complements our analysis. The work by Scabini and Bruno \cite{scabini2023structure}, demonstrating the correlation between neuronal centrality and network performance, resonates with our focus on the layer-wise node interaction attributes of analysis. Rauber et al.~\cite{rauber2016visualizing} consider dimensionality reduction of weights, of which we use a similar method in this paper to visualize weights on high dimensional space. Similarly, Naitzat et al.'s~\cite{naitzat2020topology} examination of how networks transform data topology and Semenova et al.’s investigation into noise propagation in NNs~\cite{semenova2022understanding} align with our interest in the underlying mechanisms of data representation and processing in neural models. Our approach similarly seeks to uncover the structural determinants that underlie these phenomena. 


These pivotal studies collectively advance our comprehension of DNNs, informing both theoretical insights and practical applications in the field. Our work, while building on these foundations, provides a unique perspective by emphasizing the importance of weight matrices in understanding and optimizing neural network behavior.

\section{Characterization of learnable parameters}\label{sec:method}
We present a novel methodology to systematically investigate the weight (learnable parameters) of trained NNs to uncover their learning dynamics. In our methodology, we experimented (with various trials from 30 to 1000 for each configuration depending on computational budget) with various architectures of three different deep learning models: DNNs, CNNs, and ViTs. Several architecture configurations (from very minimal to large networks) of these three models were trained on three well-known and widely used benchmark datasets of varying complexities: MNIST, Fashion MNIST, and CIFAR-10. In the first analysis stage, we focus on the weight statistics of trained networks. We computed the mean \(\mu_w = \nicefrac{1}{N} \sum_{i=1}^{N} w_i \) and standard deviation \(\sigma_w= \sqrt{\nicefrac{1}{N} \sum_{i=1}^{N} (w_i - \mu_w)^2}\) of converged \( N \) number of network weights   
\( w_i \). In addition to weight statistics to characterize the optimal and suboptimal networks, we analyze the distribution of the weights using a normalized histogram and kernel density estimation of the network weights.  

In the second stage, we focus on node strength and pair-wise node strength analysis. We, therefore, compute the strength of the nodes, which is the sum of incoming absolute weight values at a node as \(\mathrm{S} = \sum_{i=1}^{N_j} |w_i|,\),  where \(N_j\) is the total number of weight \(w_i\) incident on \(j\)-th node in a network. In the CNN case, node strength was kernels, and for ViT, it was the attention layer.

In our final stage, we project the network weights (of layers) to high-dimensional space using t-SNE~\cite{van2008visualizing} in order to assess the position of the high-dimensional weight vectors for the characterization network varied learning capabilities.

\subsection{Elements of model architecture for characterization}
When experimenting with DNN models, we selected the weight matrices between fully connected (FC) layers of DNNs, including the weight matrices between the final FC layer and the decision layer (output layer). This was to perform a layer-wise assessment of trends and variability in converged DNN weight and not only the entire network. This was done to observe closely the layer-wise differences between optimal and suboptimal networks. We vary the architectures (an input layer, two hidden layers, and an output layer) of DNNs by changing the size of their hidden layer from a very minimum network size to a large network size. This was done incrementally and systematically by increasing the network size until the performance of DNNs reached a saturation level with maximum accuracy on respective datasets (e.g., 99\%+ for the MNIST dataset)~\cite{lecun1998gradient}. 

Similarly, we use a simple architecture for CNN models with one convolutional layer and one FC layer~\cite{lecun1998gradient,simard2003best}. Their size increases similarly to the size of DNN models. We started with a minimal network and reached a maximum network when the performance on the data sets reached saturation. We also designed a minimal ViT model with an encoder that takes a minimum of 2 heads to a maximum of 16 heads~\cite{vaswani2017attention,dosovitskiy2020image}. 

We systematically increased the architecture size from a minimal architecture to an architecture that gave a high accuracy (e.g., 99\%+ for the MNIST dataset) with an aim to identify the minimal network that may perform well for these datasets and to keep the computational overhead reasonable for expensive trails. 

\subsection{Characterization of deep learning convergence profiles}
We aimed to perform the training of deep learning models over a fixed number of epochs and asses various convergence profiles. We analyze these convergence profiles and identify three groups.  The first group consisted of high-performing clusters whose accuracy was close to that of the best-performing network among all trials. The second group was low-accuracy convergence profiles whose network accuracy was close to worst-performing networks. We also identified mid-accuracy clusters of networks whose performance was close to 50\% accuracy. 
 
\subsection{Experimental setup}
\begin{table}[h]
\caption{Experiment settings for three deep learning models: DNN, CNN, and Vision Transformer (ViT). $\eta$ indicates the learning rate, C indicates the number of input channels, and $\theta$ indicates weight initialization. }\label{tab:models_settings}
\centering
\setlength{\tabcolsep}{2pt}
\begin{tabular}{cccccccc}
\hline
Network & Data & Layer & Input & Output & Activation & Setting & Value \\
\toprule
\multirow{6}{*}{DNN}  & \multirow{3}{*}{\makecell{MNIST/ \\ F-MNIST}} 
      & FC1 & 28$\times$28 & 5-200 & relu & batch & 100 \\
   &  & FC2 & 5-200 & 5-200 & relu & epochs & 20 \\
   &  & FC3 & 5-200 & 10 & softmax & optimizer & Adam \\
\cmidrule{2-8}
     & \multirow{3}{*}{CIFAR-10}
     & FC1 & 3$\times$32$\times$32 & 5-1000 & relu & $\eta$ & 0.001 \\
   & & FC2 & 5-1000 & 5-1000 & relu & $\theta$ & normal \\
   & & FC3 & 5-1000 & 10 & softmax & runs & 1000 \\
\midrule

\multirow{4}{*}{CNN} & \multirow{2}{*}{\makecell{MNIST \\ F-MNIST}} 
    & Conv1 & 28$\times$28$\times$1 & 26$\times$26$\times$C & relu & batch & 100 \\
 &  & FC & 26$\times$26$\times$C & 10 & softmax & epochs & 20 \\
\cmidrule{2-8}
    & \multirow{2}{*}{CIFAR-10}           
    & Conv1 & 3$\times$32$\times$32 & 30$\times$30$\times$C & relu & $\eta$ & 0.001 \\
  & & FC & 30$\times$30$\times$C & 10 & softmax & $\theta$ & normal \\
  & &  &  &  &  & runs & 1000 \\
\midrule

\multirow{3}{*}{ViT} & \multirow{3}{*}{\makecell{MNIST \\ F-MNIST \\ CIFAR-10}} 
      & Encoder & \makecell{d\_model=784 \\ nhead=2-16} & - & - & batch & 100 \\
    & & FC & 784 (input) & 10 & - & $\eta$ & 0.001 \\
    & & - & - & - & - & runs & 30 \\
\bottomrule
\end{tabular}
\end{table}
In our research, we adopted three computer vision datasets: MNIST, Fashion MNIST (FMNIST), and CIFAR-10. These datasets are widely used in deep learning and image classification, providing diversity and complexity (difficulty level) for our deep learning network analysis. Table~\ref{tab:models_settings} shows the complete set of experiments for DNN, CNN, and ViT models.  

We experimented with DNNs as initial networks on image classification tasks. The DNN model architecture includes an input layer, two hidden layers, and an output layer. These layers are connected through \textit{relu} activation function. The input size corresponds to the image size of the respective datasets, and the output size is 10, which corresponds to the number of categories for classification. Our model also includes two hidden layers, and the number of nodes in each hidden layer varies between 5 and 200 to achieve feature representation. 

To ascertain the ubiquity of the observed phenomena beyond DNN networks, experiments were extended to both CNNs and Transformer-based architectures (ViTs). Table~\ref{tab:models_settings} shows the network and hyperparameters settings. Our CNN model consists of a convolutional layer and an FC followed by a global average pooling layer. The vision transformer, which contains an encoder with a subsequent fully connected layer, was used. 

The specific parameter settings are as follows: For the training, the batch size is 100, epochs are set to 20, and the optimizer used is Adam. The learning rate $\eta$ was set to 0.001, and the network weight was initialized using a normal distribution. The total number of experiments conducted was 1000 for DNNs and CNNs experiments. However, for ViT experiments, due to computational costs, 30 trials were performed across all datasets. The consistent choice of hyperparameters across the datasets ensures a standardized experimental setup, enabling a direct comparison of results across these datasets.

These hyperparameters were chosen to ensure consistent training across different datasets while still allowing efficient convergence. Adam optimizer was chosen for its adaptive learning rate capabilities and normal weight initialization, which ensure a symmetrical distribution of weights at the start of training. Running the experiment 1000 times provides a comprehensive understanding of the model's performance across different initializations and random shuffles.

\section{Results and analysis}\label{sec:results}
We initially conducted a series of 1000 runs of experiments on three datasets using DNN/CNN networks and 30 runs of experiments of ViTs detailed in Table~\ref{tab:models_settings}. The aggregated results are summarized in Table~\ref{tab:results} and show substantial variations (uncertainty) in training accuracy under both identical and diverse network configurations. Such disparities are visible across all three datasets: MNIST, FMNIST, and CIFAR-10. This indicates that similar network architecture/training setups yield a wide spectrum of performances. In Table~\ref{tab:results}, results were grouped into three accuracy groups: low accuracy, mid accuracy, and high accuracy.
\begin{table}
\centering
\caption{Accuracy distribution groups (low, mid, and high) of varied deep learning models (DNNs, CNNs, and ViTs) for their several experiment runs as mentioned in Table~\ref{tab:models_settings} over three datasets. MNIST, Fashion MNIST (F-MNIST), and CIFAR-10. In Table~\ref{tab:results}, min, med, and max indicate the minimum, median, and maximum accuracy values of the group. The group 'non' indicates the group of models that do not converge.}
\label{tab:results}
\setlength{\tabcolsep}{3.5pt}
\begin{tabular}{cccccccccccc}
\hline
   & \multicolumn{4}{c}{MNIST} & \multicolumn{3}{c}{F-MNIST} & \multicolumn{3}{c}{CIFAR-10} \\ \midrule
network & group & min & med & max &  min & med & max &  min & med & max \\ \midrule
\multirow{4}{*}{DNN}
 & non & - & 11.35 & - &  - & 10.00 & 30.00 & - & 11.35 & 20.00 \\ 
 & low & 30.00 & 39.33 & 55.00  & 30.00 & 72.90 & 75.00 & 20.00 & 30.94 & 32.00 \\ 
 & mid & 80.00 & 80.91 & 82.00  & 83.50 & 83.67 & 84.00 & 45.00 & 52.29 & 55.00 \\ 
 & high & 95.00 & 98.55 & 100.00 & 95.00 & 95.88 & 100.00 & 75.00 & 77.80 & 100.00 \\ \midrule
\multirow{3}{*}{CNN}
 & low & 0.00 & 93.77 & 95.00 & 0.00 & 85.93 & 90.00 &  0.00 & 40.39 & 55.00 \\ 
 & mid & 96.00 & 97.68 & 98.00   & 96.00 & 97.71 & 98.00 &  55.00 & 63.36 & 75.00 \\ 
 & high & 99.50 & 99.87 & 100.00 & 99.50 & 99.95 & 100.00 & 80.00 & 88.91 & 100.00 \\ \midrule
\multirow{3}{*}{ViT}
 & low & 0.00 & 22.11 & 30.00 & 0.00 & 30.40 & 40.00 & 0.00 & 17.50 & 20.00 \\ 
 & mid & 70.00 & 74.58 & 85.00 & 65.00 & 69.59 & 72.00 &  0.00 & 38.18 & 40.00 \\ 
 & high & 85.00 & 90.71 & 100.00 & 74.00 & 74.59 & 100.00 &  40.00 & 44.56 & 100.00 \\ 
 \bottomrule
\end{tabular}
\end{table}

\begin{figure}
    \centering
    \setlength{\tabcolsep}{1pt}
    \begin{tabular}{cccc}
    \label{fig:convergence}
    \centering
         & MNIST & FMNIST & CIFAR-10\\
        \rotatebox[origin=c]{90}{DNN}
            & \raisebox{-0.5\height}{\includegraphics[width=0.33\textwidth]{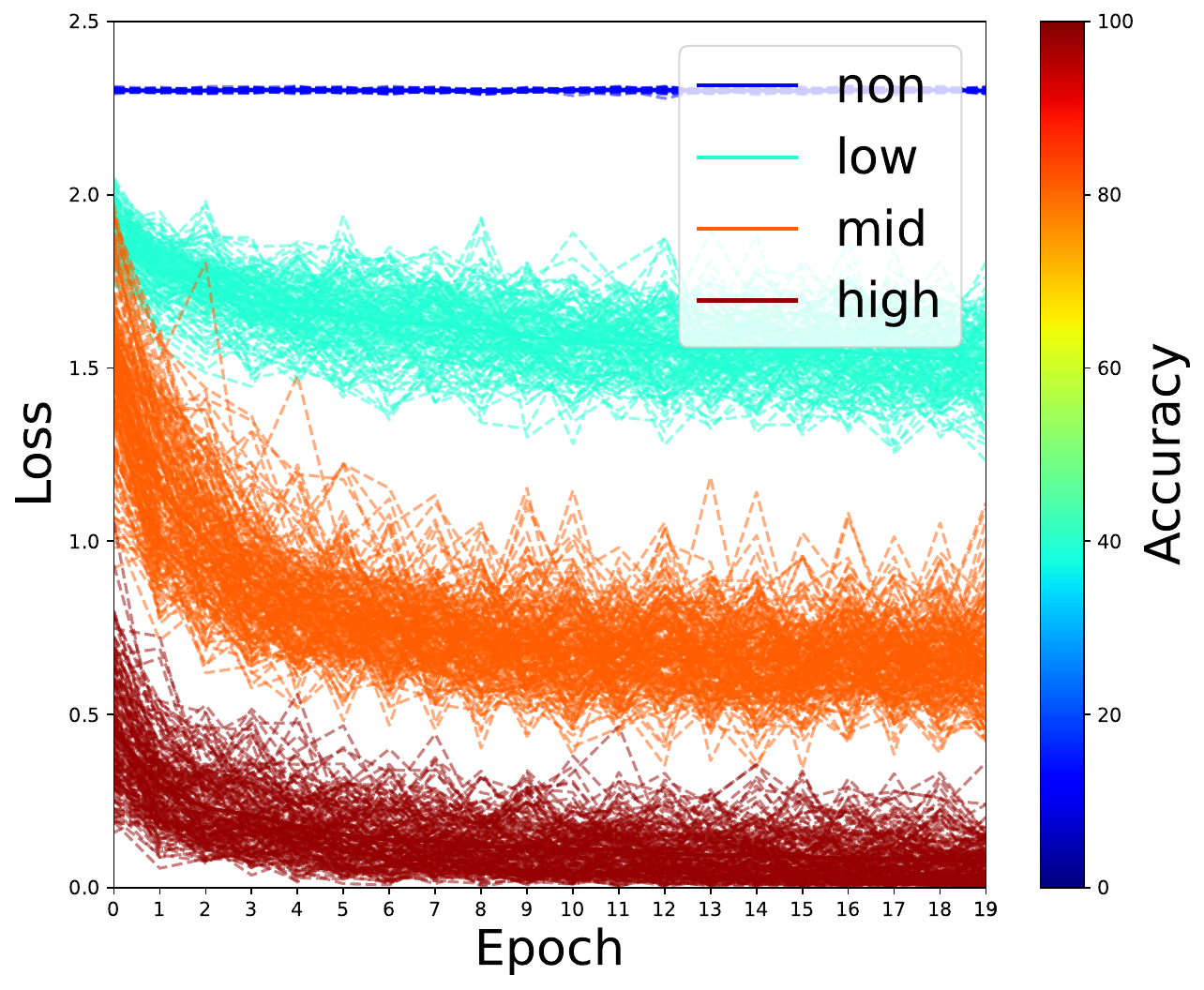}} 
            & \raisebox{-0.5\height}{\includegraphics[width=0.33\textwidth]{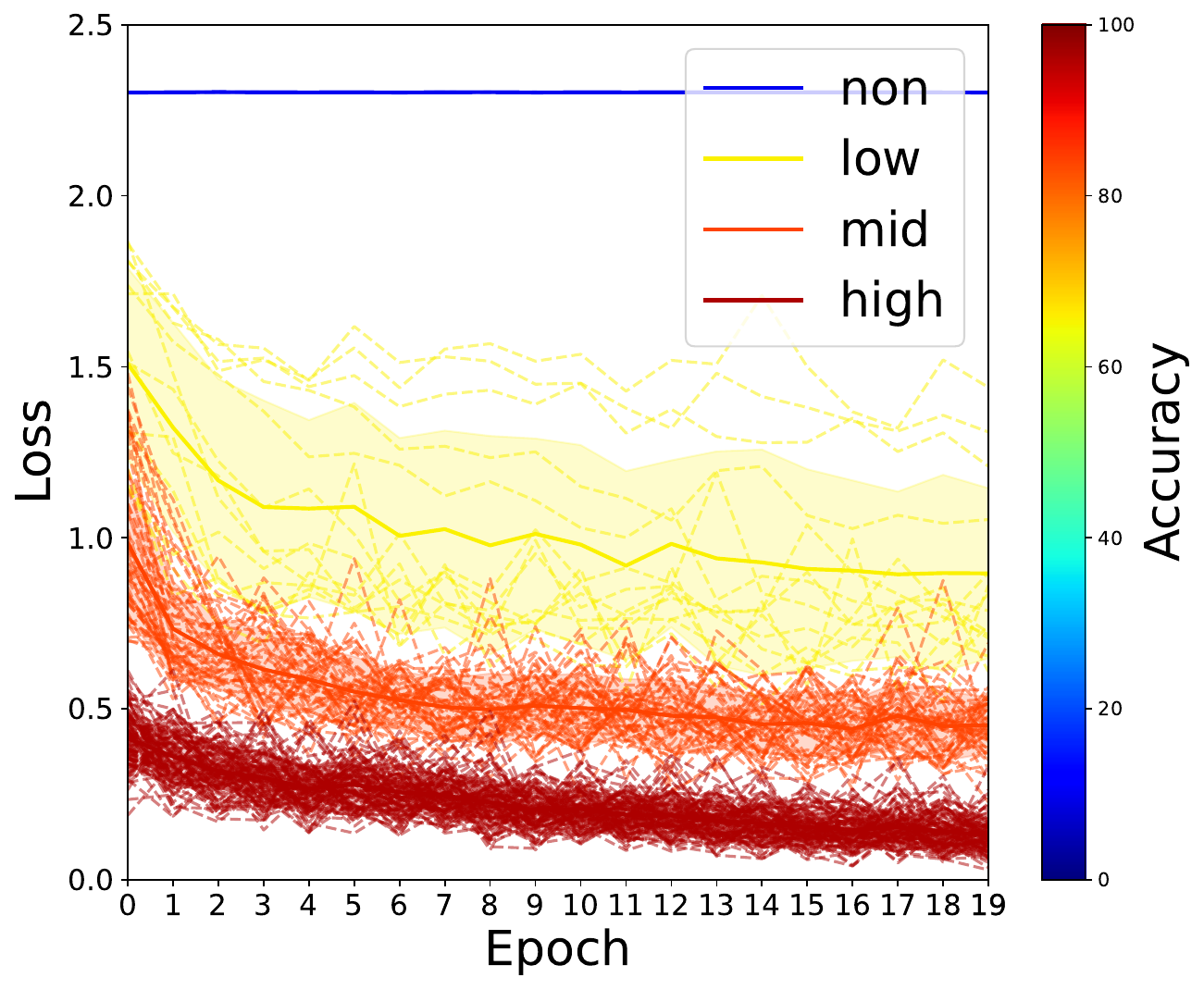}}
            & \raisebox{-0.5\height}{\includegraphics[width=0.33\textwidth]{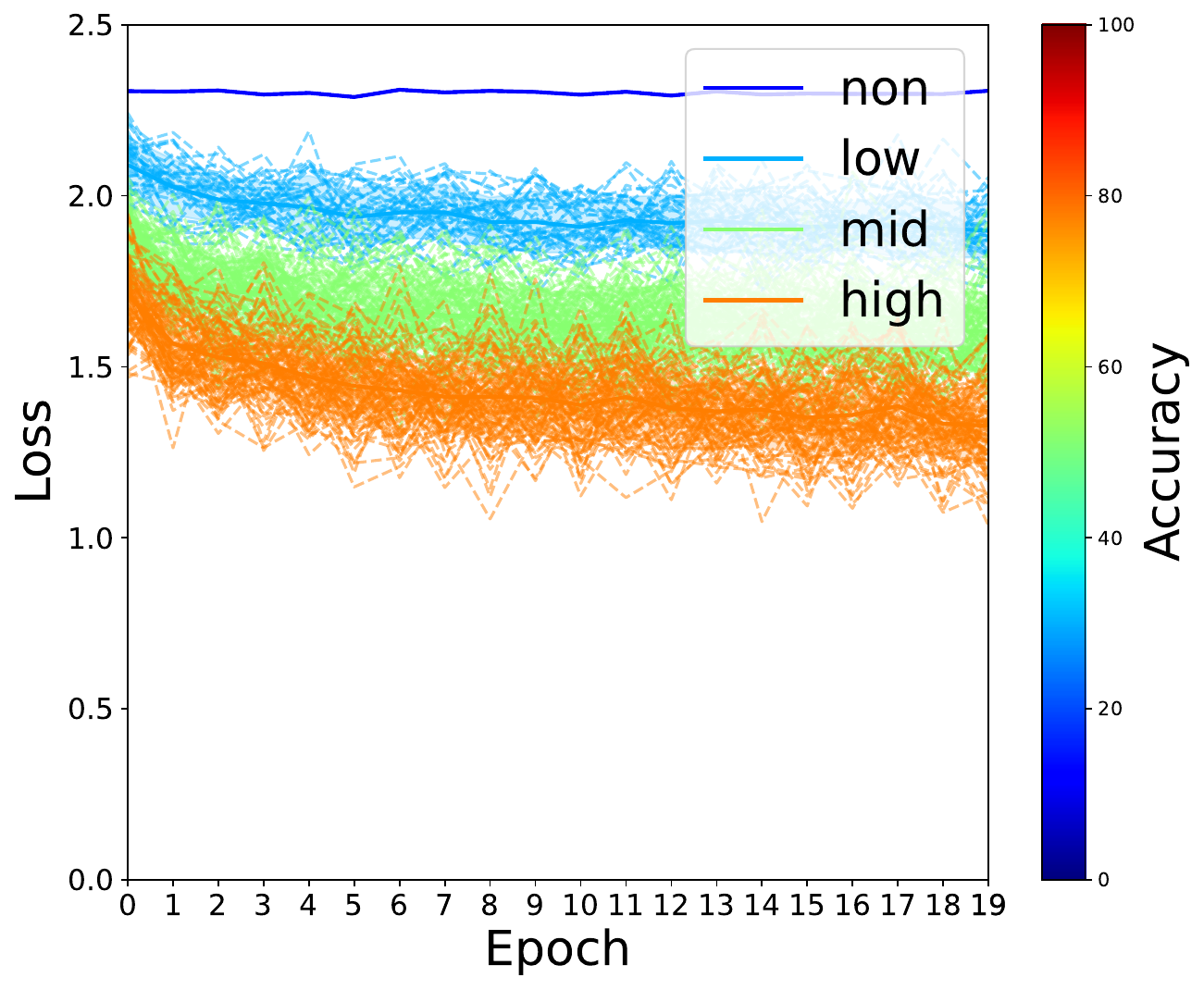}} \\
         \rotatebox[origin=c]{90}{CNN} 
            & \raisebox{-0.5\height}{\includegraphics[width=0.33\textwidth]{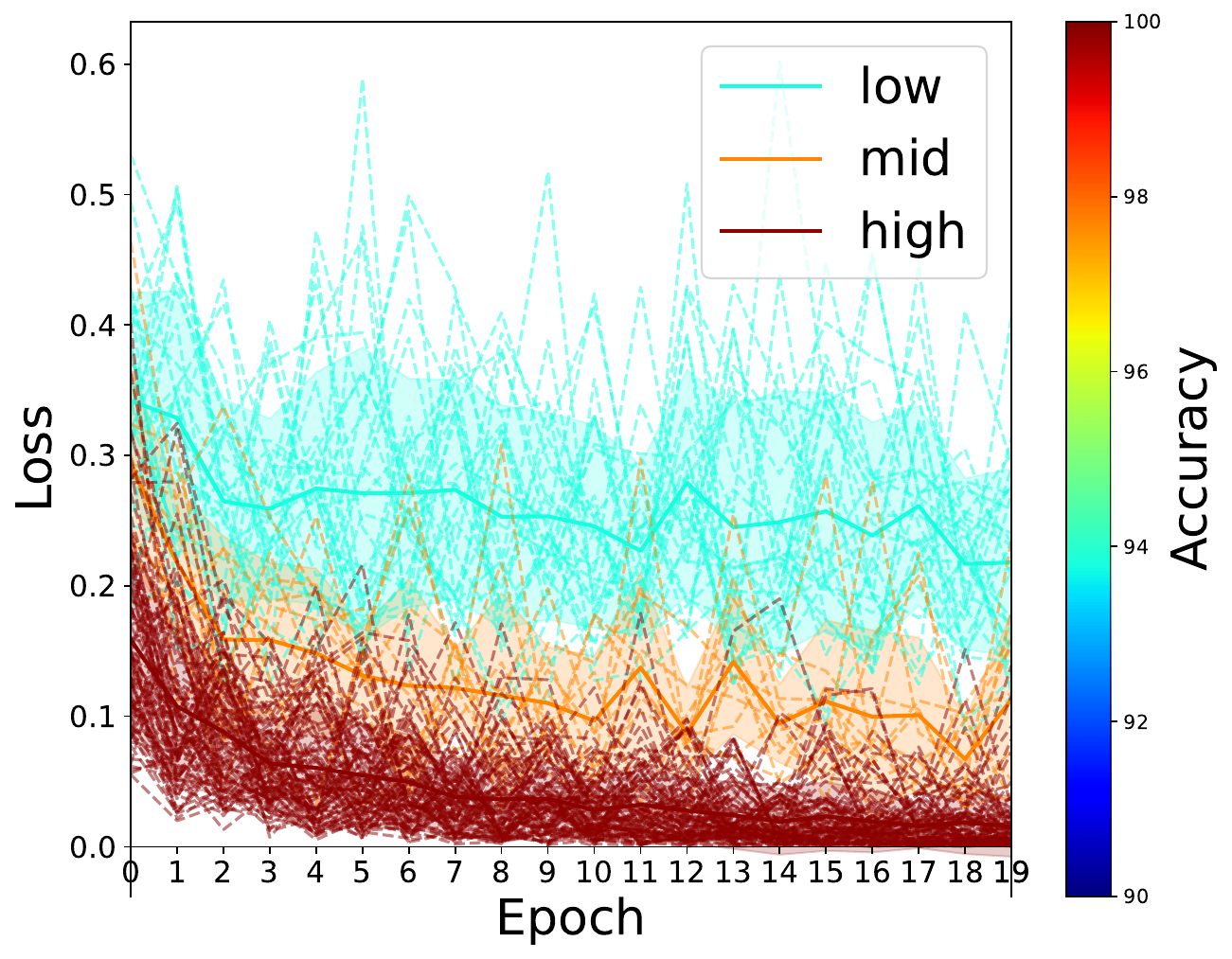}} 
            & \raisebox{-0.5\height}{\includegraphics[width=0.33\textwidth]{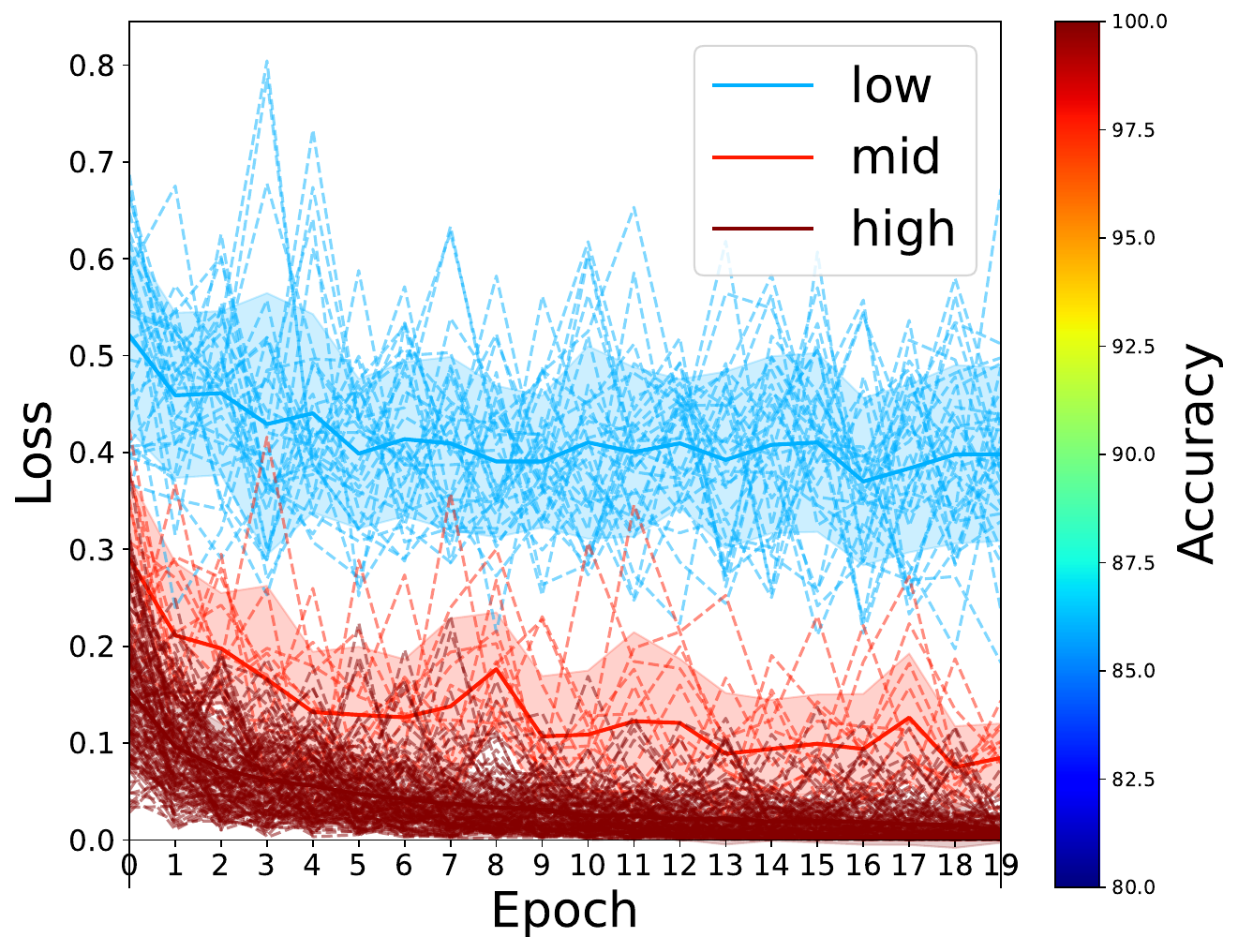}} 
            & \raisebox{-0.5\height}{\includegraphics[width=0.33\textwidth]{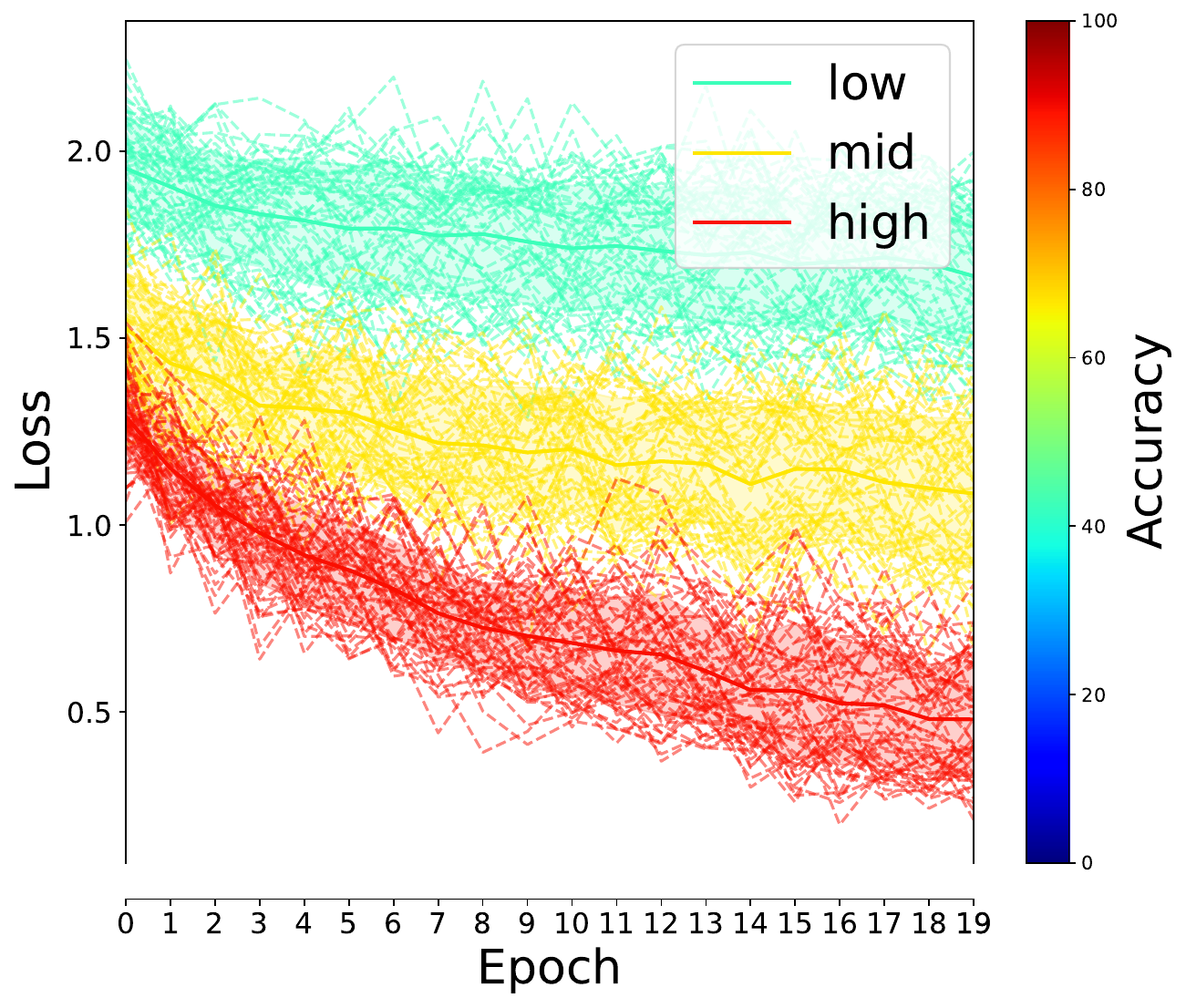}} \\
        \rotatebox[origin=C]{90}{ViT} 
            & \raisebox{-0.5\height}{\includegraphics[width=0.33\textwidth]{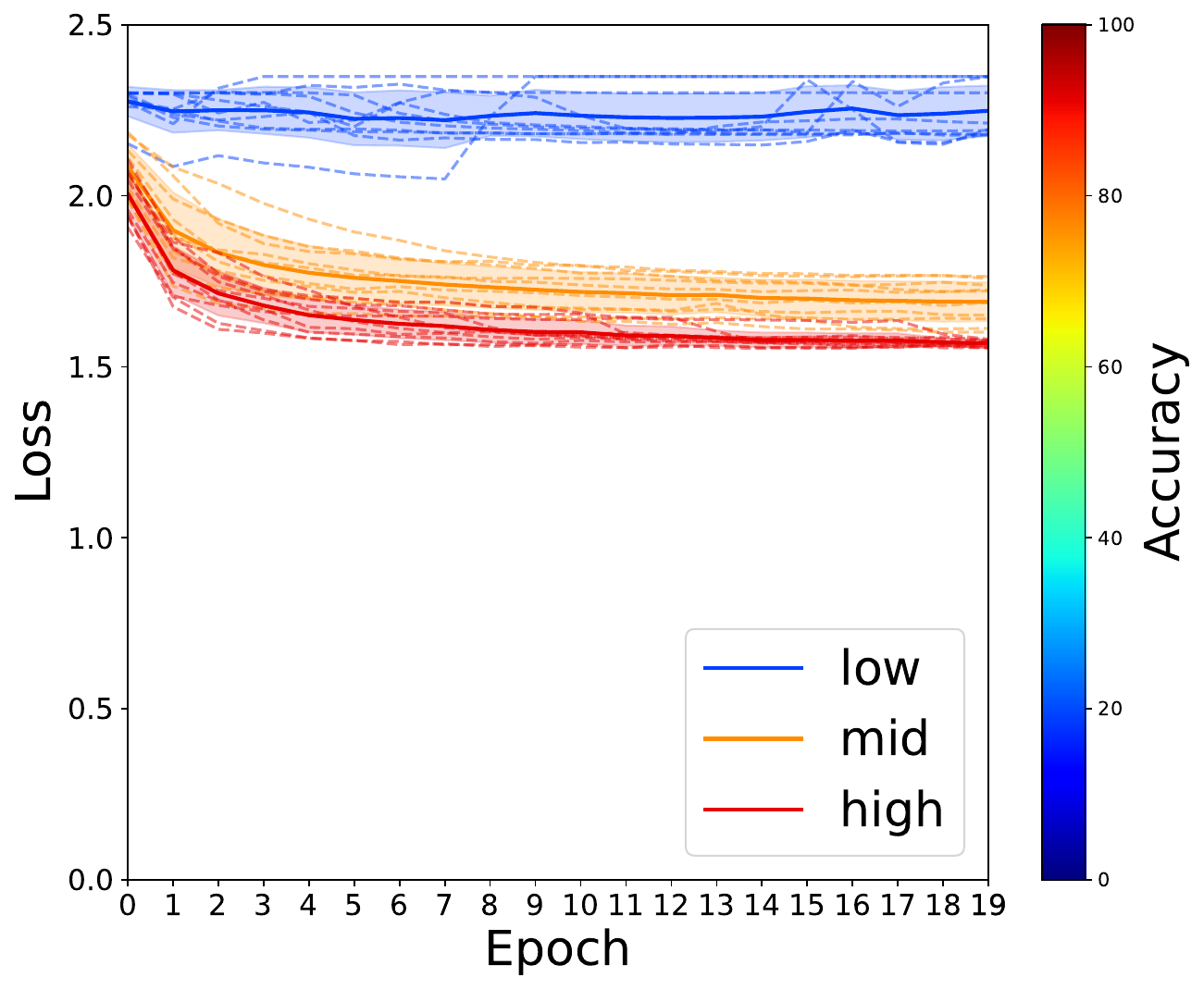}} 
            & \raisebox{-0.5\height}{\includegraphics[width=0.33\textwidth]{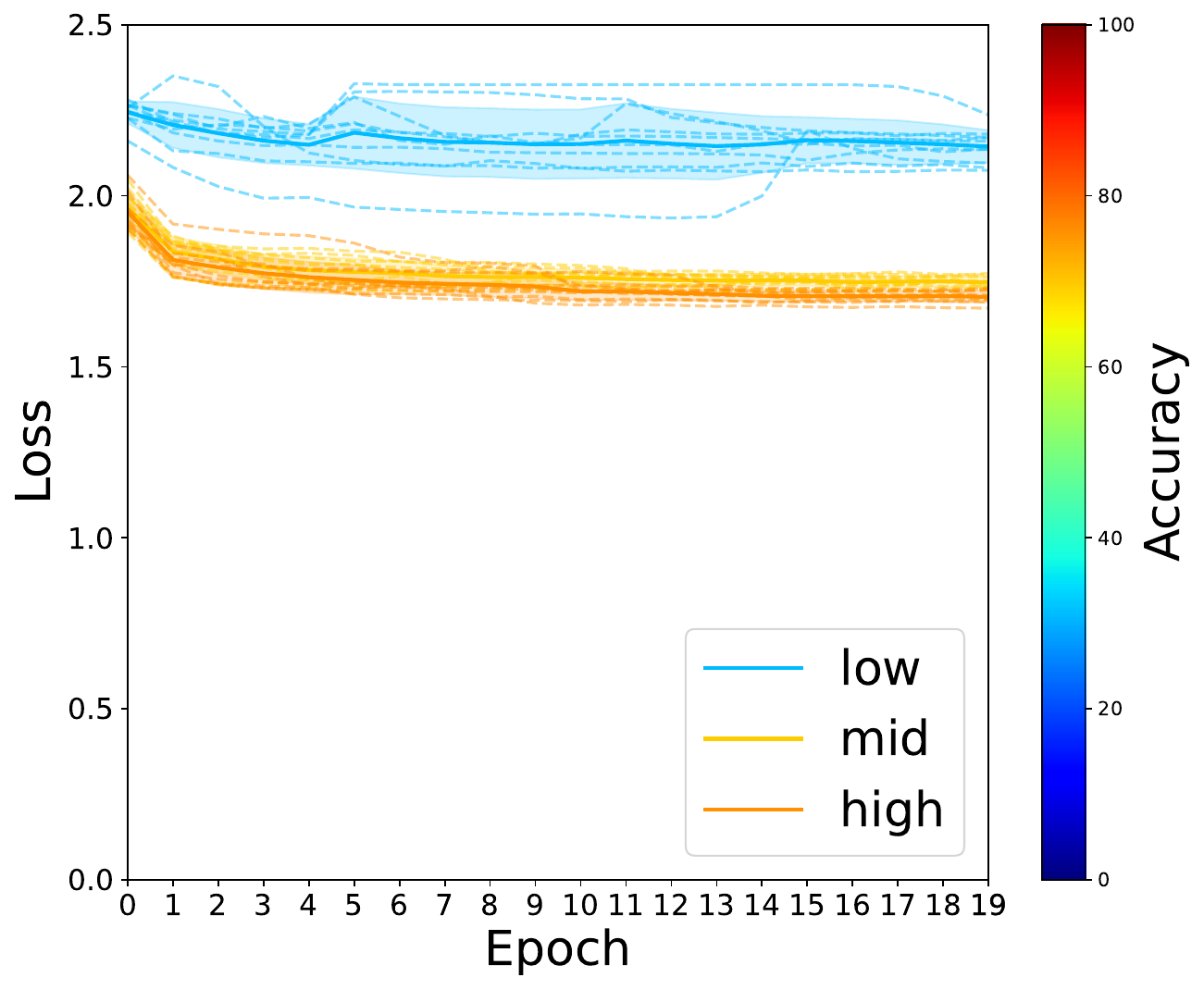}}
            & \raisebox{-0.5\height}{\includegraphics[width=0.33\textwidth]{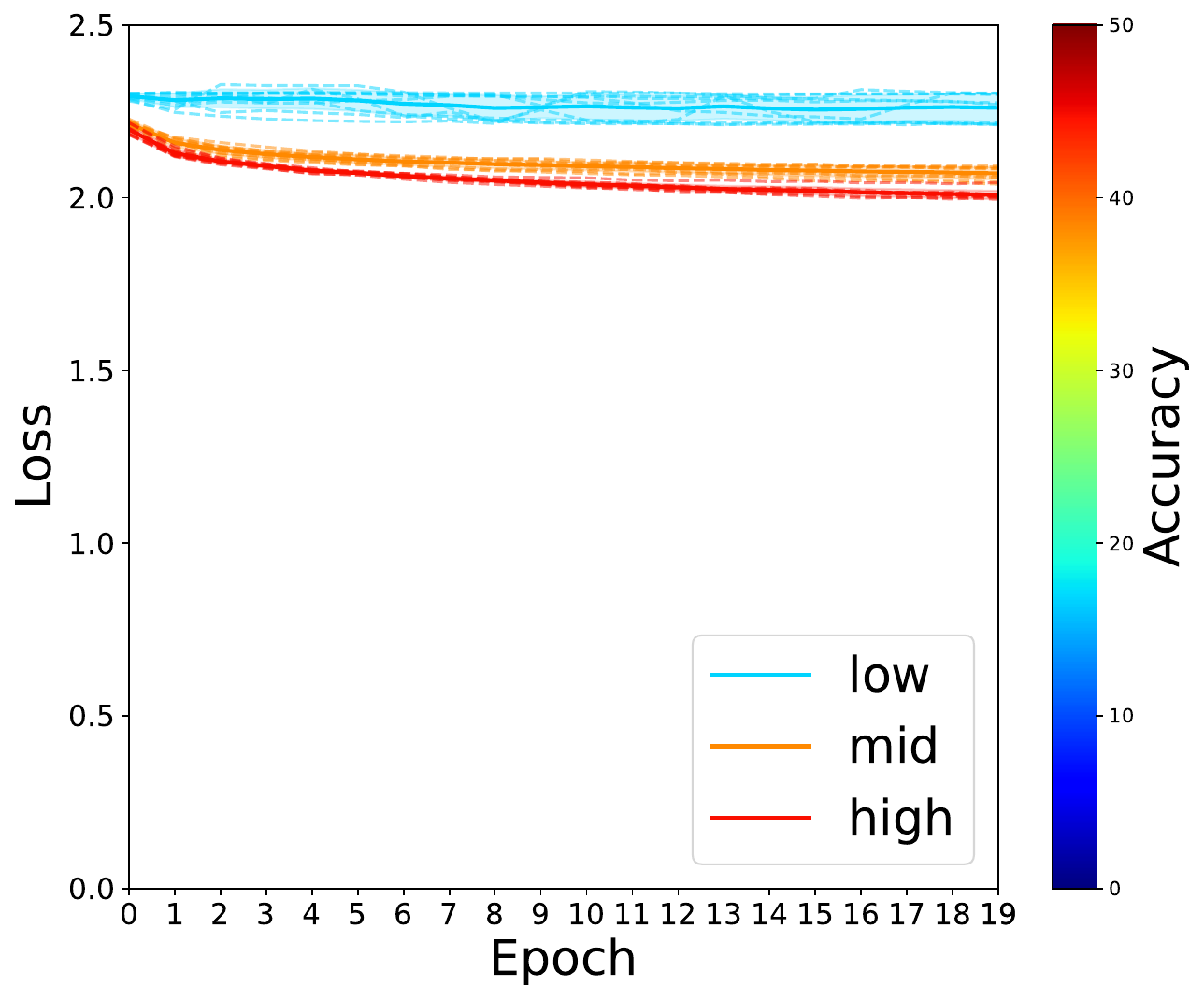}} \\
    \end{tabular}
    \caption{Convergence of various DNN, CNN, and ViT models over MNIST, FMNIST, and CIFAR-10 datasets. The x-axis and y-axis represent the training epoch and training loss. The color represents the test accuracy of models. Models convergence profiles are categorized into low, mid, and high accuracy model groups. For DNN, the group `non' (in blue) indicates a group of models that happened to be initialized randomly around zero weight and consequently have not converged over epochs due to fluctuation in gradient around zero and vanishing of gradients to previous layers and were no part of the subsequent analysis. The model characterization is therefore performed for successful networks (high accuracy), marginally successful networks (mid accuracy), and failed networks (low accuracy).
    }
\end{figure}

Observing the convergence of networks, as shown in Fig.~\ref{fig:convergence}, when training proceeds, network losses that reach higher accuracy steadily and continuously decrease, indicating that learning is effective. There is a sharp contrast in the convergence behavior between high- and low-accuracy networks - the former group shows a centralized approach that minimizes the loss, while the later group of lower-accuracy networks fluctuates greatly, with the convergence plot fluctuating for signs that show the struggle to extract and preserve patterns critical to high performance. 

For DNN experiments, there is a group called `non,' which are the networks that did not converge. These networks are also visible in blue color convergence lines in Fig.~\ref{fig:convergence}. We thoroughly analyzed these non-converging networks and found that these networks were the minimal `input-5-5-class-category' DNN architecture whose weight initialization was close to zero, and their computed gradient fluctuated in both directions and did not propagate beyond the output layer in any of the 20 epochs of the training.

The experimental results in Table~\ref{tab:results} show considerable variety in the accuracy of networks, whether using DNN, CNN, or ViT. This high degree of uncertainty in performance may be due to various reasons, including random weight initialization, random shuffling of training data, and the optimization algorithm used. However, the training settings were similar to our experiments. Thus, this variety indicates real-world deep learning model training challenges, and the results demonstrate that training uncertainty and fluctuations are pervasive in network structure. This is not just a problem with a specific network or initialization method but a common deep-learning phenomenon.

\begin{figure}    
    \centering
    \fontsize{7pt}{7pt}\selectfont
    \setlength{\tabcolsep}{5pt}
    \centering 
       \begin{tabular}{ll}
        & \begin{tabular}{cccc}
             ~~~\qquad I/P - FC1 & ~\quad~\qquad FC1 - FC2 & ~\quad~\qquad FC2 -O/P & ~\quad~\quad Whole Net \\
        \end{tabular} \\
        \rotatebox[origin=c]{90}{\centering MNIST} & \raisebox{-0.5\height}{\includegraphics[width=0.95\textwidth]{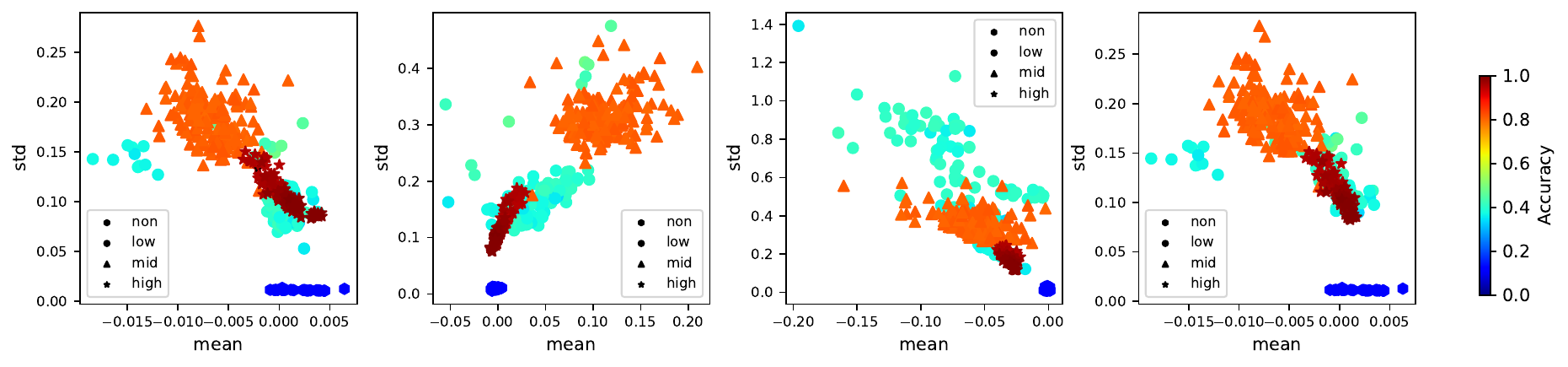}} \\
        \rotatebox[origin=c]{90}{FMNIST} & \raisebox{-0.5\height}{\includegraphics[width=0.95\textwidth]{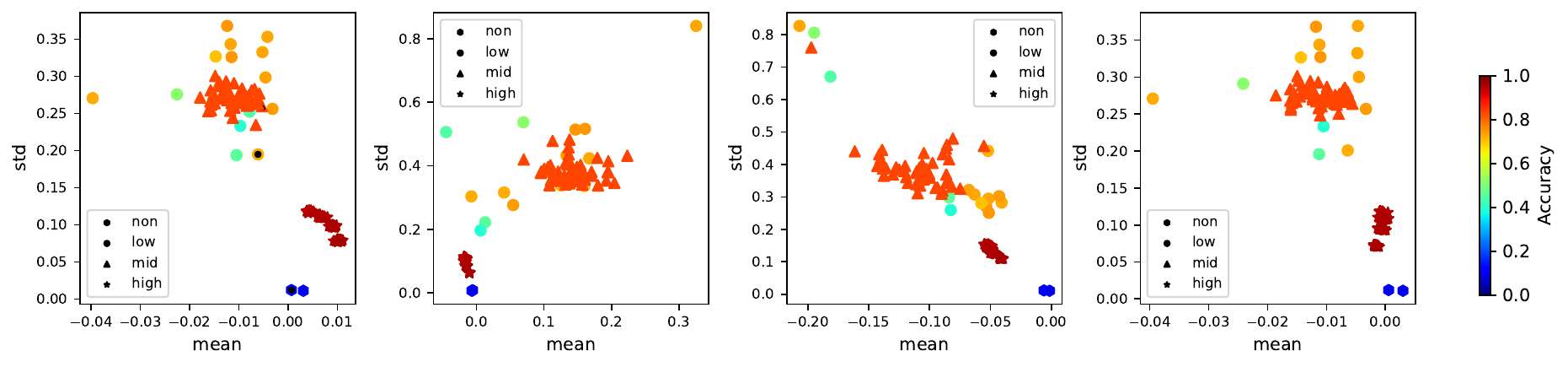}} \\
        \rotatebox[origin=c]{90}{CIFAR-10} & \raisebox{-0.5\height}{\includegraphics[width=0.95\textwidth]{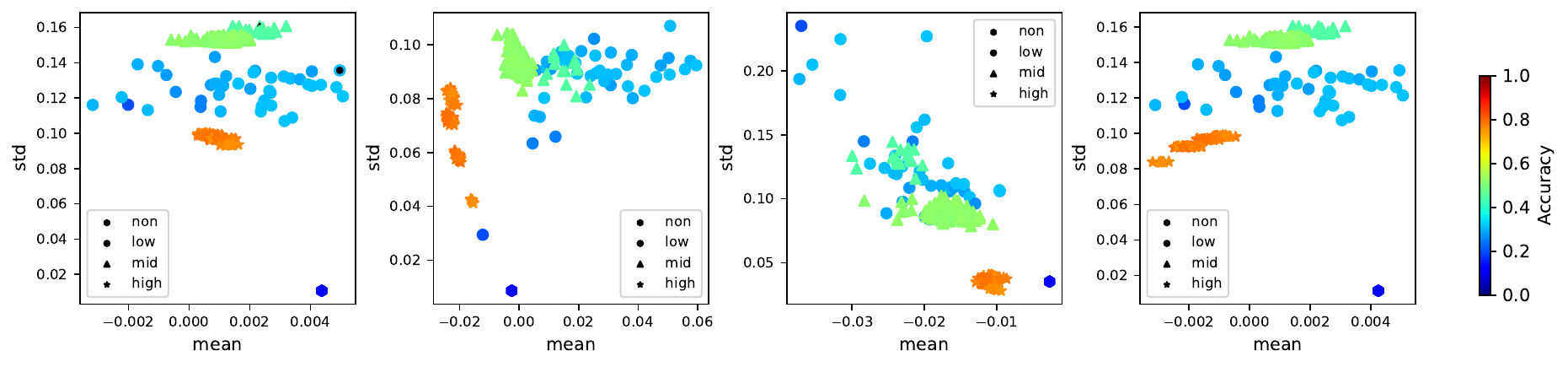}} \\
        \end{tabular}  \\ 
    \caption{DNN weight analysis for optimal and suboptimal network characterization.   
    }
    \label{fig:dnn_weight_meanstd}
\end{figure}

\begin{figure}    
    \centering
    \fontsize{7pt}{7pt}\selectfont
    \setlength{\tabcolsep}{5pt}
    \centering 
       \begin{tabular}{ll}
        & \begin{tabular}{cccc}
             ~\quad~\qquad Conv1 & ~\quad~\quad~\quad\qquad ~~FC & ~\quad~\quad~\qquad\qquad All &  \\
        \end{tabular}  \\ 
        \rotatebox[origin=c]{90}{MNIST} &
           \begin{tabular}{cc} 
             \includegraphics[width=0.75\textwidth]{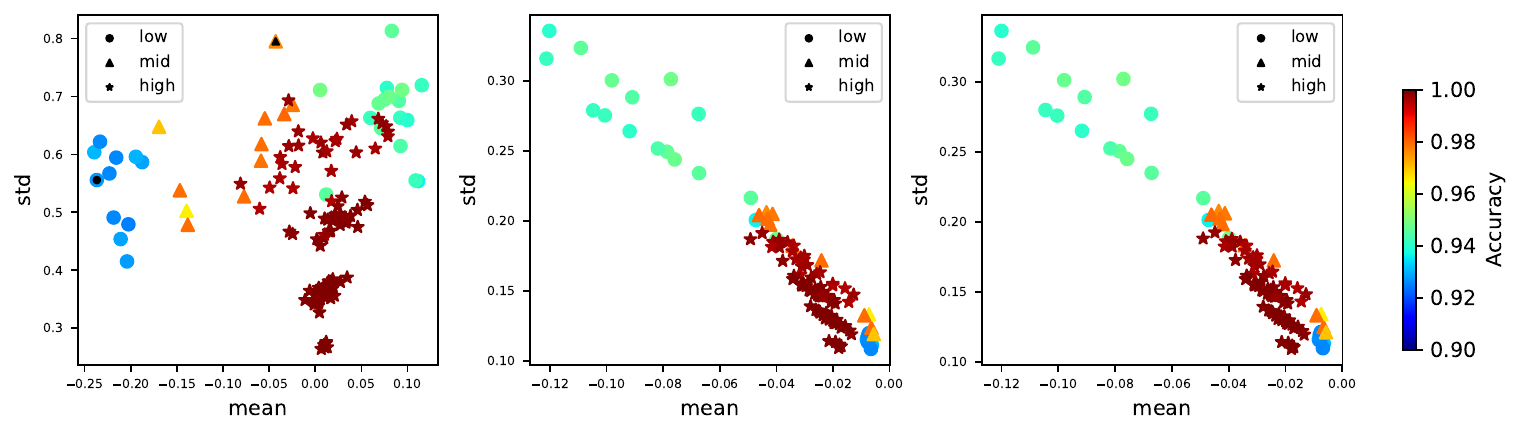} & \includegraphics[width=0.15\textwidth]{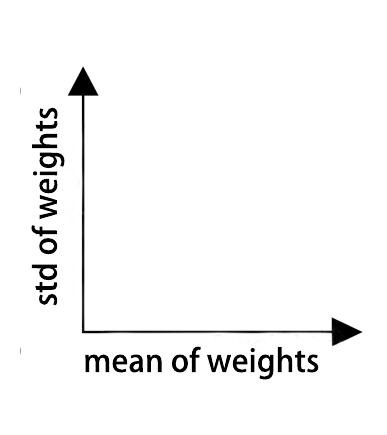}
           \end{tabular} \\ 
        \rotatebox[origin=c]{90}{FMNIST} & 
        \begin{tabular}{cc} 
             \includegraphics[width=0.75\textwidth]{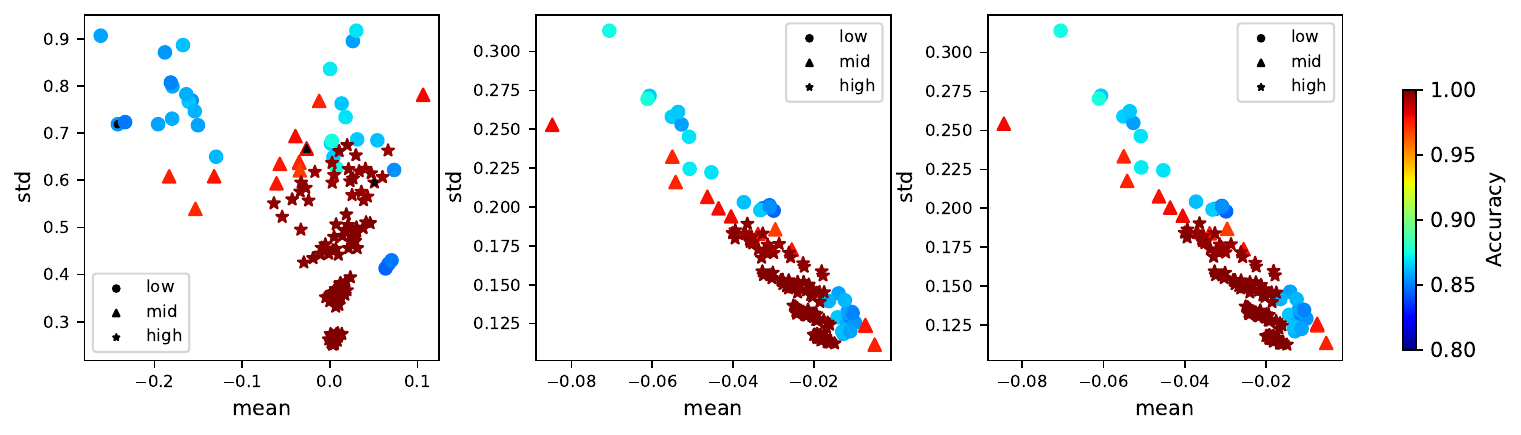} & \includegraphics[width=0.15\textwidth]{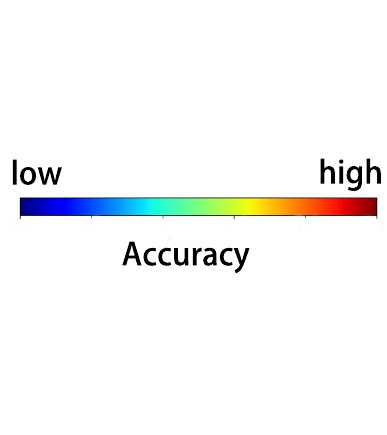}
           \end{tabular} \\ 
        \rotatebox[origin=c]{90}{CIFAR-10} & \begin{tabular}{cc} 
             \includegraphics[width=0.75\textwidth]{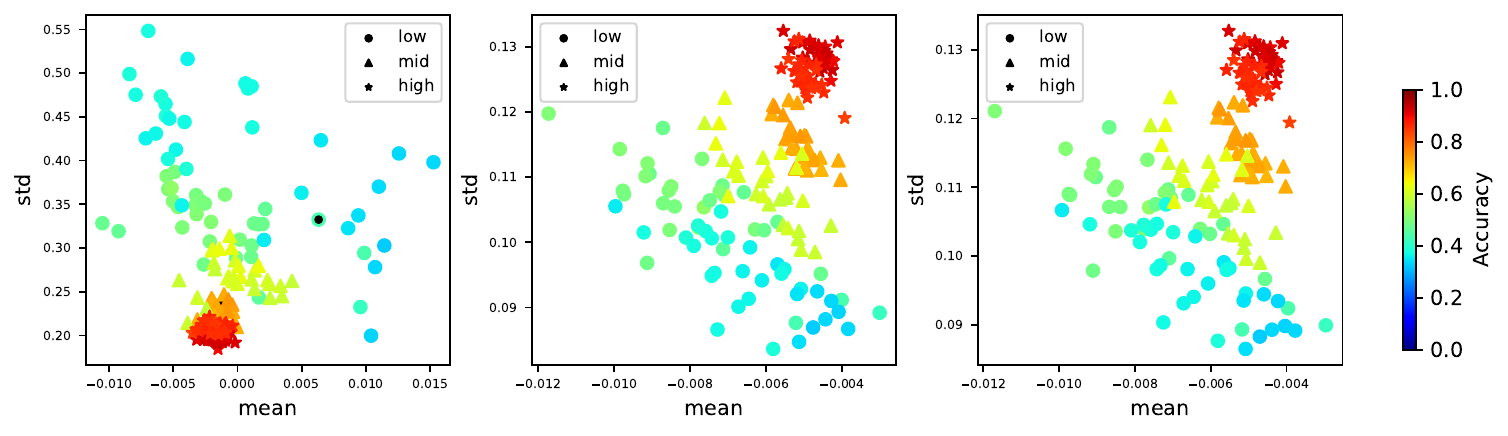} & \includegraphics[width=0.15\textwidth]{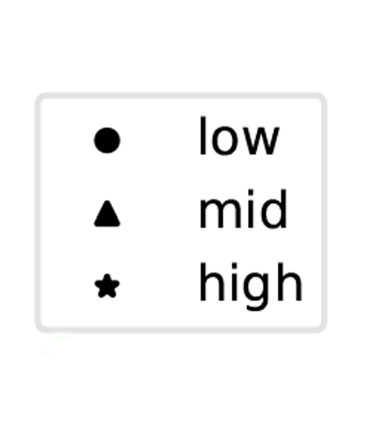}
        \end{tabular}
        \end{tabular} 
    \caption{CNN weight analysis for optimal and suboptimal networks characterization. 
    }
    \label{fig:cnn_weight_meanstd}
\end{figure}

\begin{figure}    
    \centering
    \fontsize{7pt}{7pt}\selectfont
    \setlength{\tabcolsep}{5pt}
    \centering 
        \begin{tabular}{ll}
         & \begin{tabular}{cccc}
             ~\quad~\qquad Attn & ~\quad~\qquad~\qquad MLP & ~\quad~\qquad~\qquad Norm &  ~\quad~\qquad~\qquad All \\
        \end{tabular} \\
        \rotatebox[origin=c]{90}{MNIST} & \raisebox{-0.5\height}{\includegraphics[width=0.95\textwidth]{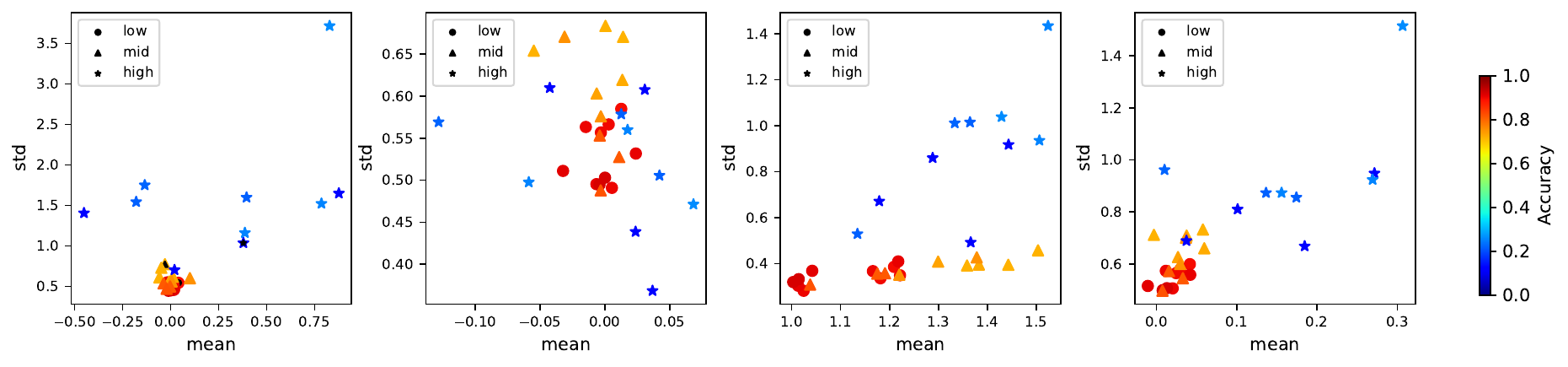}} \\
        \rotatebox[origin=c]{90}{FMNIST} & \raisebox{-0.5\height}{\includegraphics[width=0.95\textwidth]{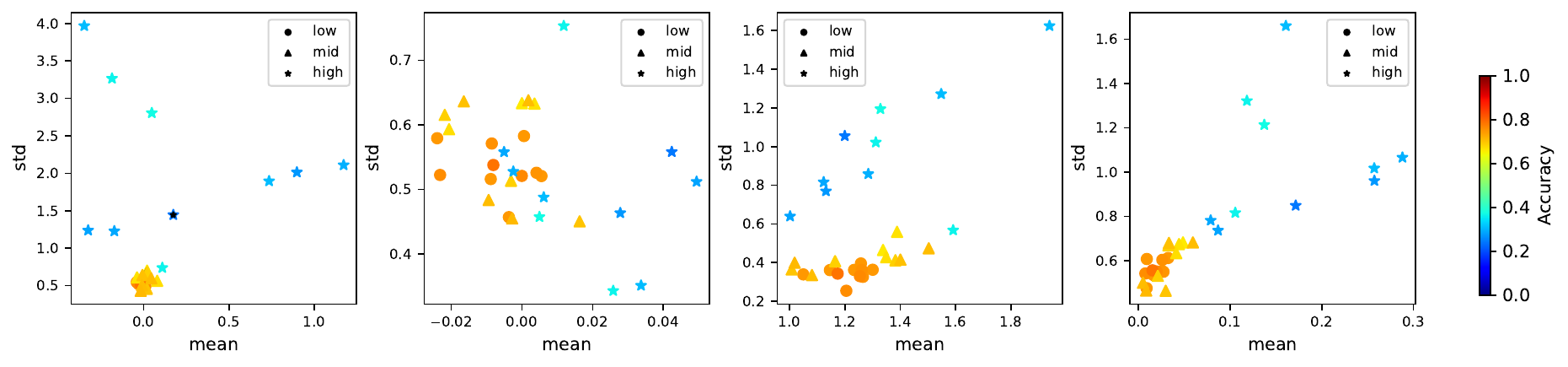}} \\
        \rotatebox[origin=c]{90}{CIFAR-10} & \raisebox{-0.5\height}{\includegraphics[width=0.95\textwidth]{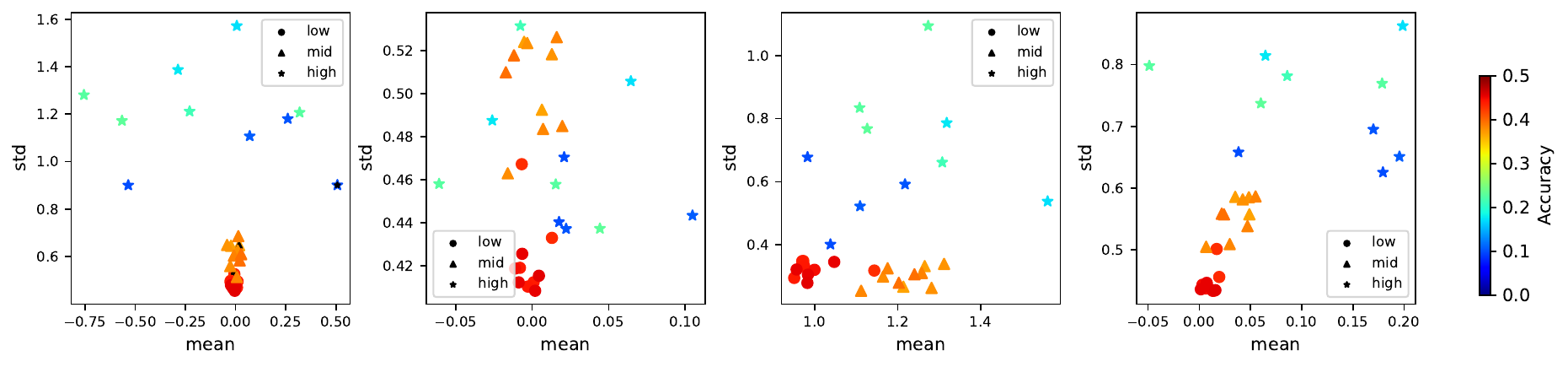}} \\
    \end{tabular}
    \caption{ViT weight analysis for optimal and suboptimal networks characterization. 
    }
    \label{fig:tf_weight_meanstd}
\end{figure}

\begin{figure}    
    \centering
    \fontsize{7pt}{7pt}\selectfont
    \setlength{\tabcolsep}{5pt}
    \centering 
       \begin{tabular}{ll}
        & \begin{tabular}{cccc}
             ~\qquad I/P - FC1 & ~~~~\qquad\quad FC1 - FC2 & ~\qquad~~~~~\quad FC2 -O/P & ~\qquad\quad~\quad Whole Net \\
        \end{tabular} \\
        \rotatebox[origin=c]{90}{\centering MNIST} & \raisebox{-0.5\height}{\includegraphics[width=0.95\textwidth]{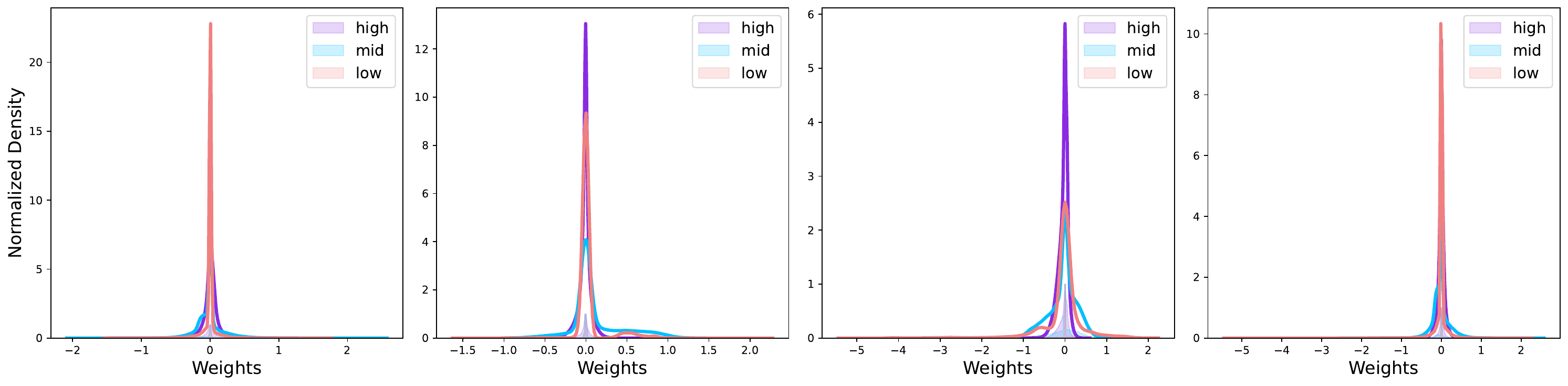}} \\
        \rotatebox[origin=c]{90}{FMNIST} & \raisebox{-0.5\height}{\includegraphics[width=0.95\textwidth]{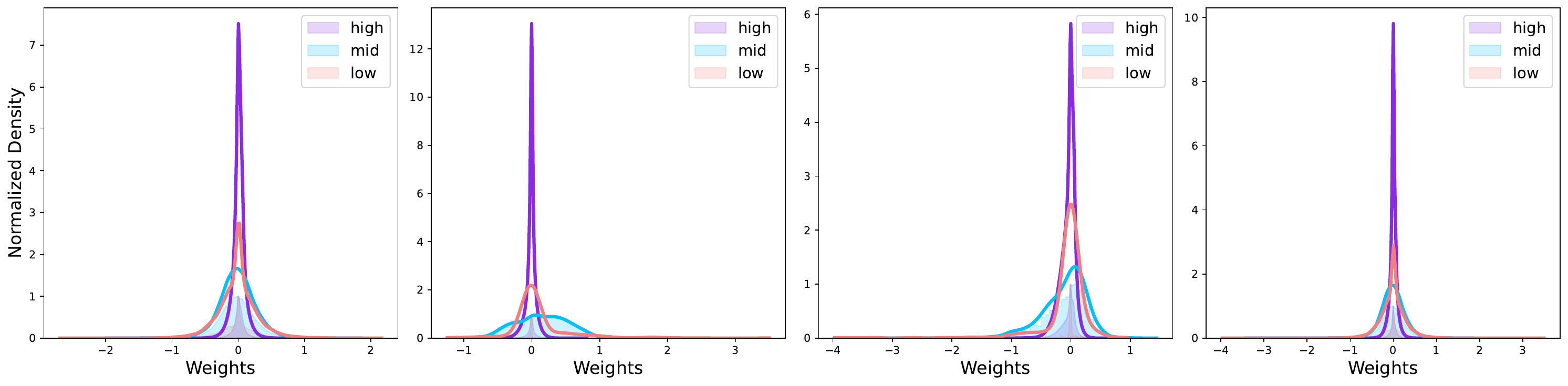}} \\
        \rotatebox[origin=c]{90}{CIFAR-10} & \raisebox{-0.5\height}{\includegraphics[width=0.95\textwidth]{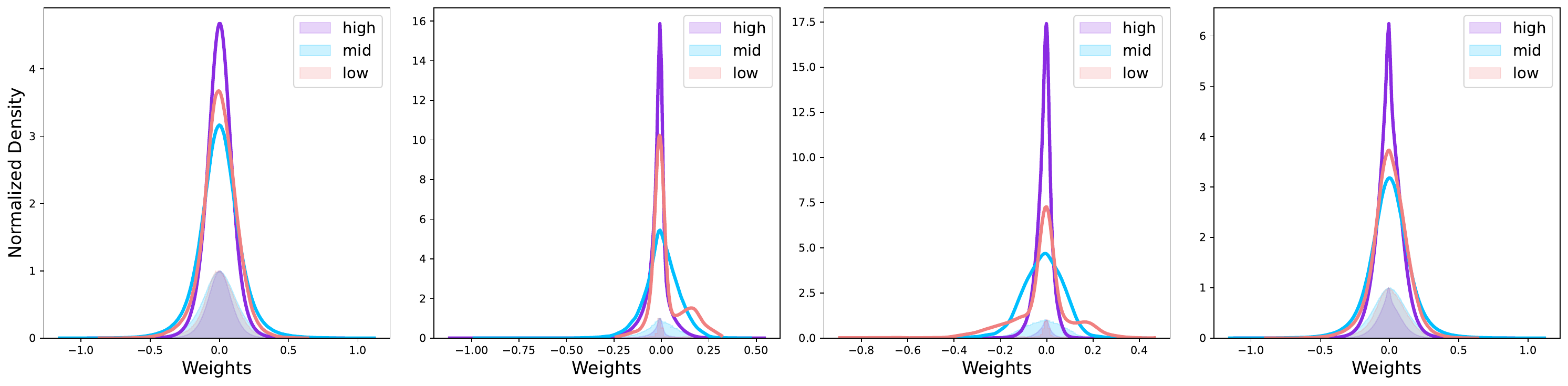}} \\
        \end{tabular}  \\ 
    \caption{DNN normalized weight distribution for model characterization. 
    }
    \label{fig:dnn_weight_distribution}
\end{figure}

\begin{figure}    
    \centering
    \fontsize{7pt}{7pt}\selectfont
    \setlength{\tabcolsep}{5pt}
    \centering 
       \begin{tabular}{ll}
        & \begin{tabular}{cccc}
            \qquad\qquad Conv1 & ~\qquad\qquad \quad~~FC & ~\qquad\qquad\qquad All &  \\
        \end{tabular}  \\ 
        \rotatebox[origin=c]{90}{MNIST} &
           \begin{tabular}{cc} 
             \includegraphics[width=0.70\textwidth]{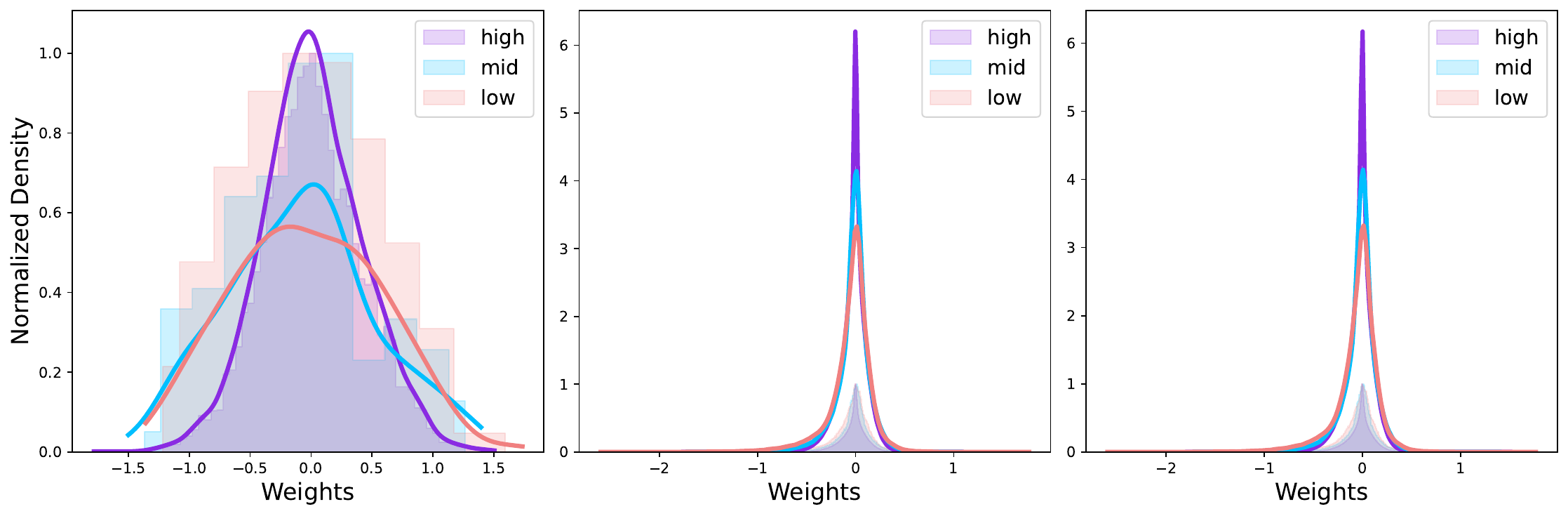} & \includegraphics[width=0.15\textwidth]{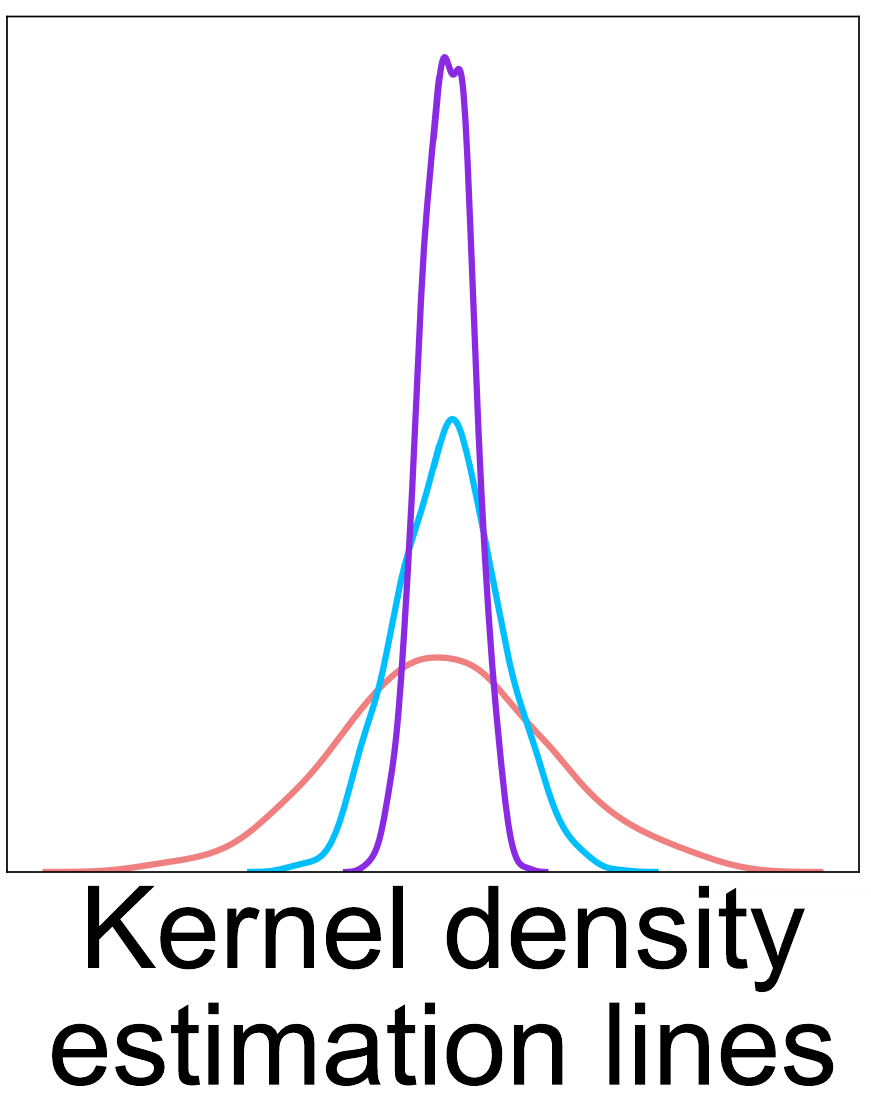}
           \end{tabular} \\ 
        \rotatebox[origin=c]{90}{FMNIST} & 
        \begin{tabular}{cc} 
             \includegraphics[width=0.70\textwidth]{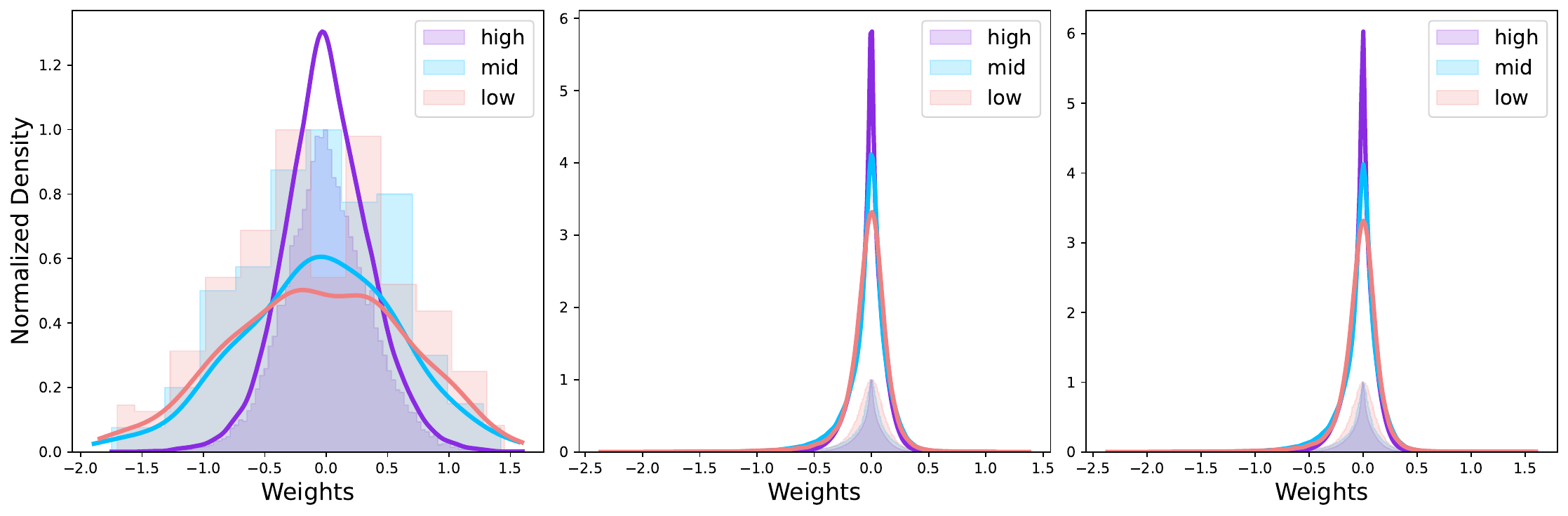} & \includegraphics[width=0.15\textwidth]{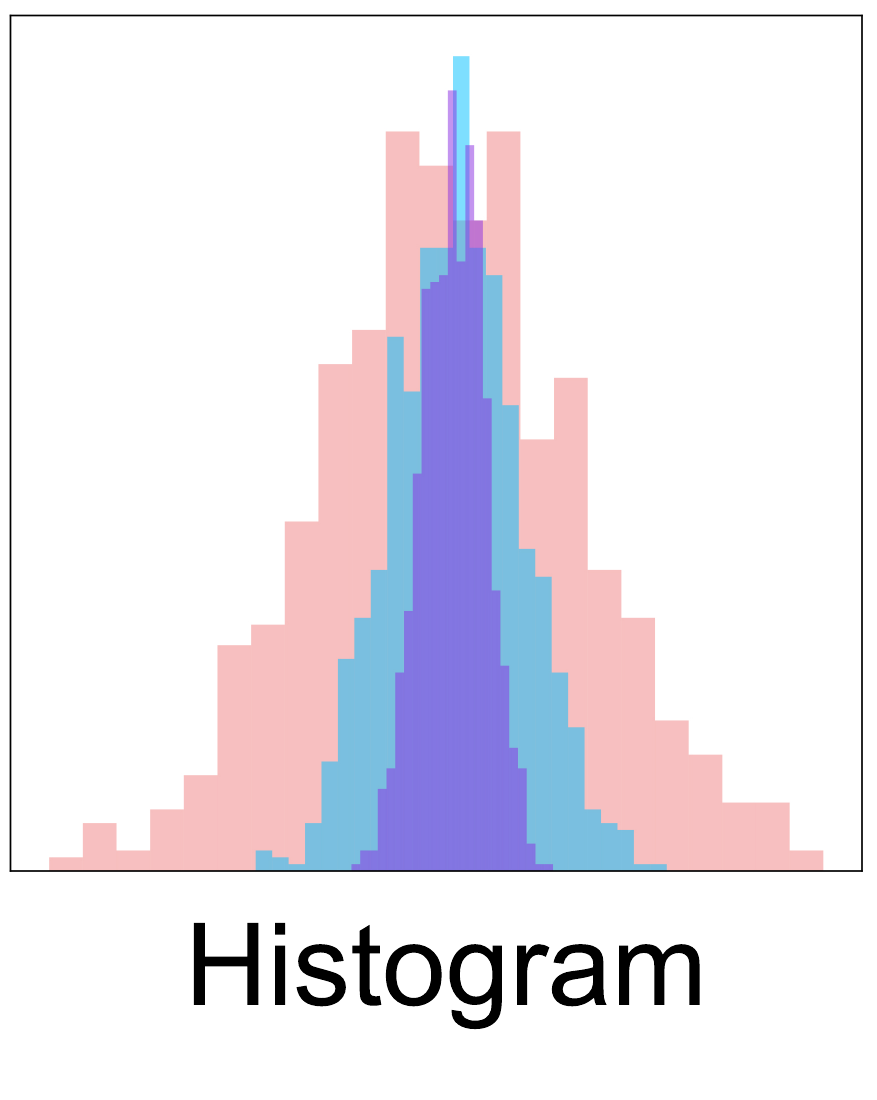}
           \end{tabular} \\ 
        \rotatebox[origin=c]{90}{CIFAR-10} & \begin{tabular}{cc} 
             \includegraphics[width=0.70\textwidth]{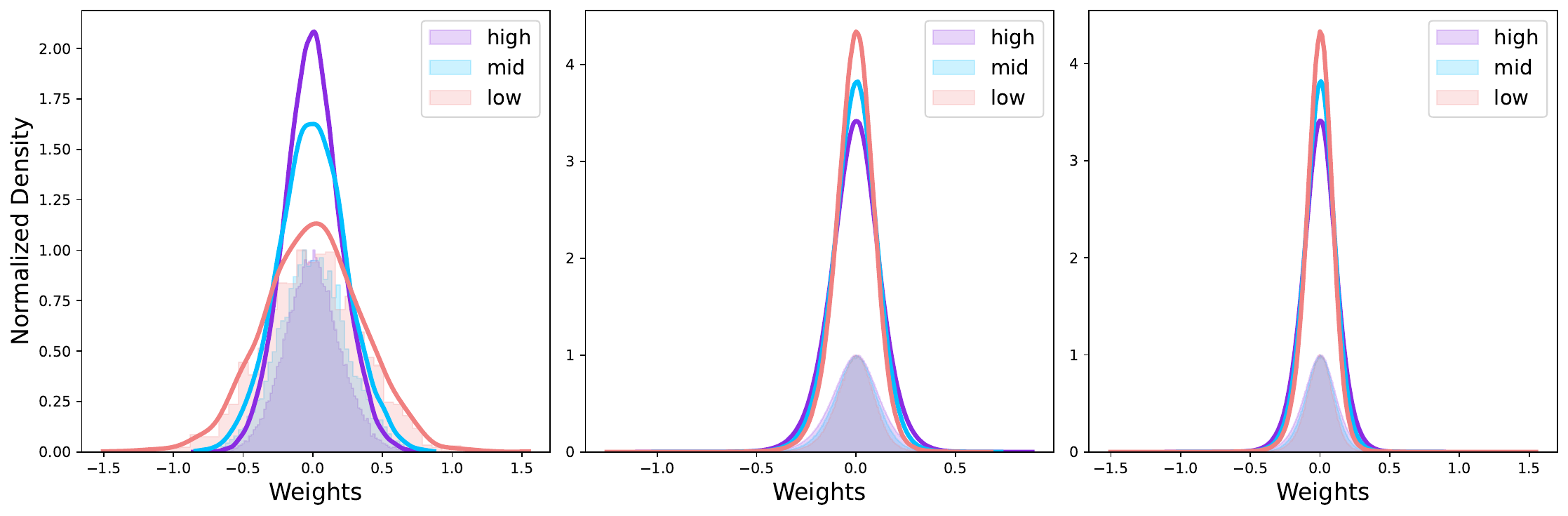} & \includegraphics[width=0.15\textwidth]{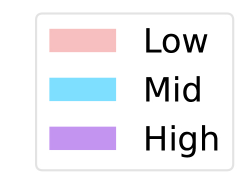}
        \end{tabular}
        \end{tabular} 
    \caption{CNN normalized weight distribution for model characterization.  
    }
    \label{fig:cnn_weight_distribution}
\end{figure}

\begin{figure}    
    \centering
    \fontsize{7pt}{7pt}\selectfont
    \setlength{\tabcolsep}{5pt}
    \centering 
        \begin{tabular}{ll}
         & \begin{tabular}{cccc}
             ~~~\qquad\quad Attn & ~~\quad\qquad\qquad MLP & \quad\qquad\qquad ~Norm & ~\qquad\qquad~ All \\
        \end{tabular} \\
        \rotatebox[origin=c]{90}{MNIST} & \raisebox{-0.5\height}{\includegraphics[width=0.85\textwidth]{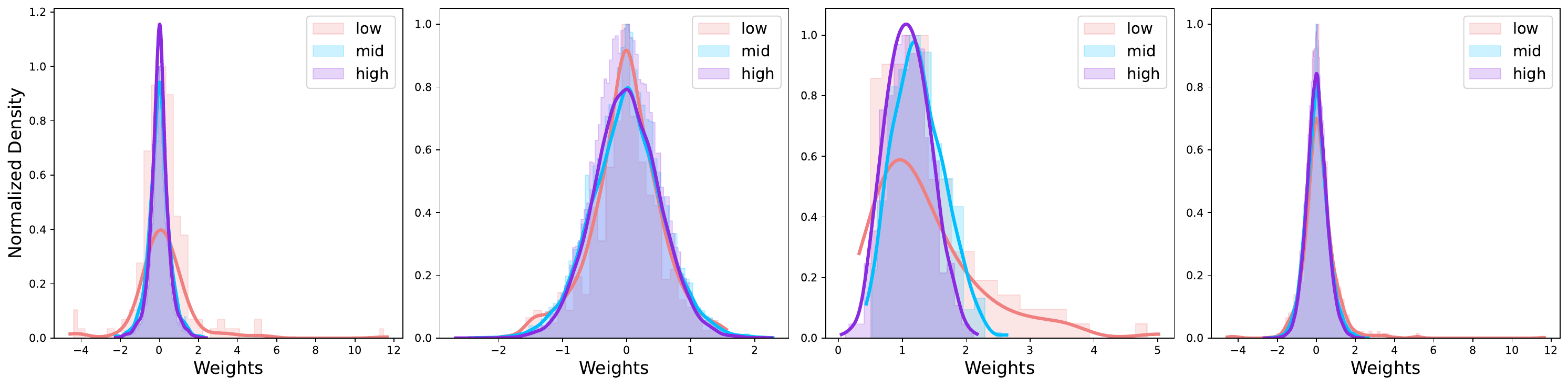}} \\
        \rotatebox[origin=c]{90}{FMNIST} & \raisebox{-0.5\height}{\includegraphics[width=0.85\textwidth]{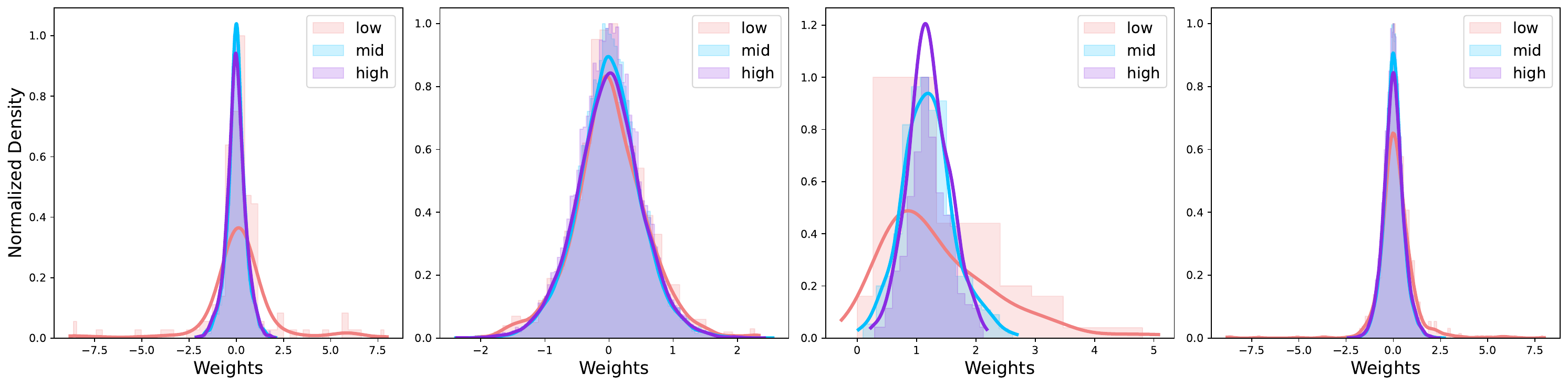}} \\
        \rotatebox[origin=c]{90}{CIFAR-10} & \raisebox{-0.5\height}{\includegraphics[width=0.85\textwidth]{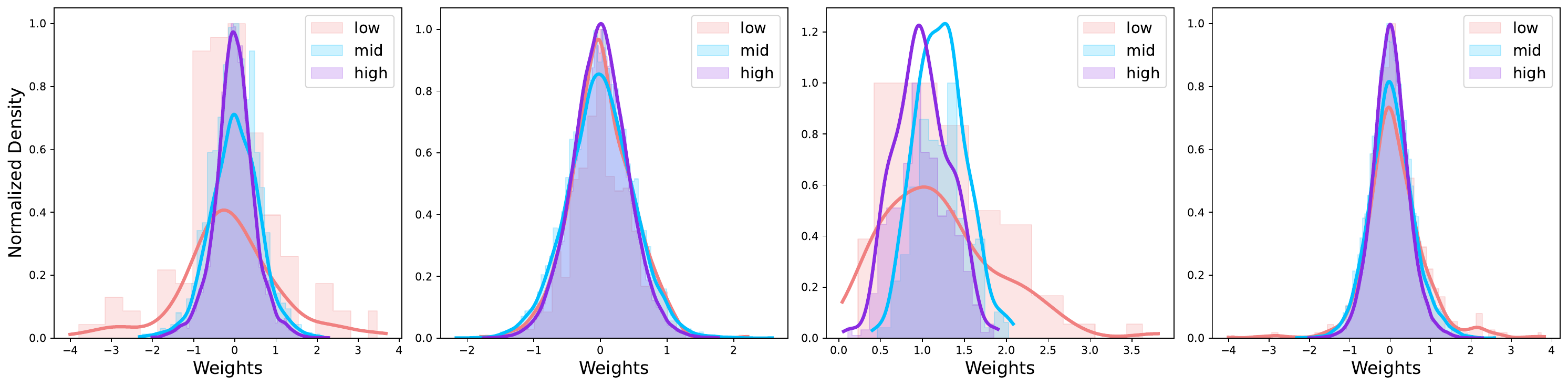}} \\
    \end{tabular}
    \caption{ViT normalized weight distribution for model characterization.  
    }
    \label{fig:tf_weight_distribution}
\end{figure}

\begin{figure}
    \centering
    \fontsize{7pt}{7pt}\selectfont
    \setlength{\tabcolsep}{5pt}
    \centering 
    \begin{tabular}{ll}
        & \begin{tabular}{cccc}
            \qquad\quad FC1 vs FC2 & \qquad\quad  FC1 vs O/P & \qquad\quad FC2 vs O/P \\
        \end{tabular} \\
        \rotatebox[origin=c]{90}{MNIST} & \raisebox{-0.5\height}{\includegraphics[width=0.80\textwidth]{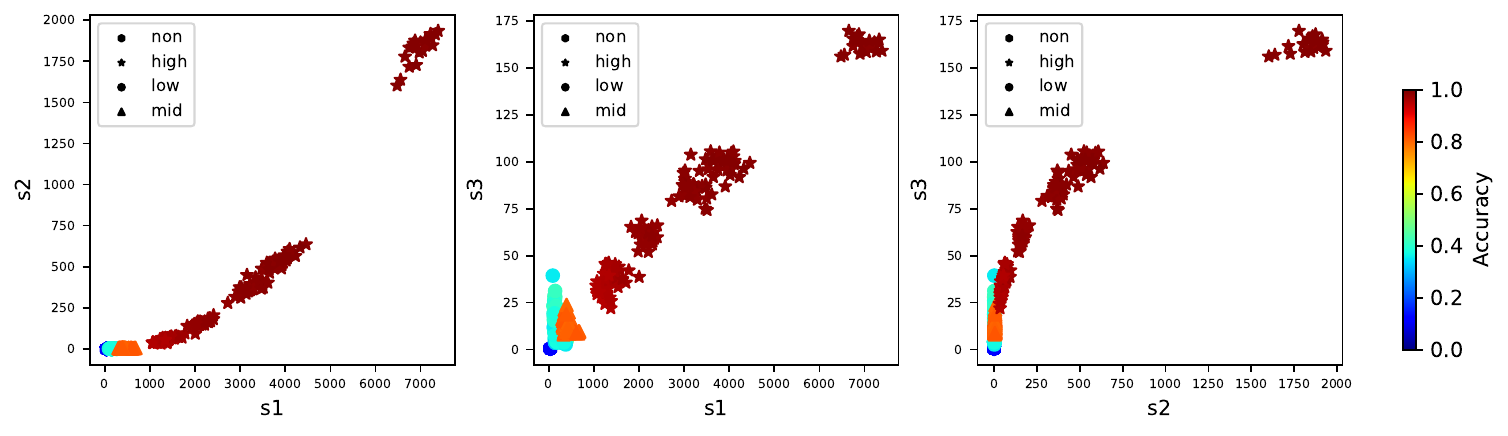}} \\
        \rotatebox[origin=c]{90}{FMNIST} & \raisebox{-0.5\height}{\includegraphics[width=0.80\textwidth]{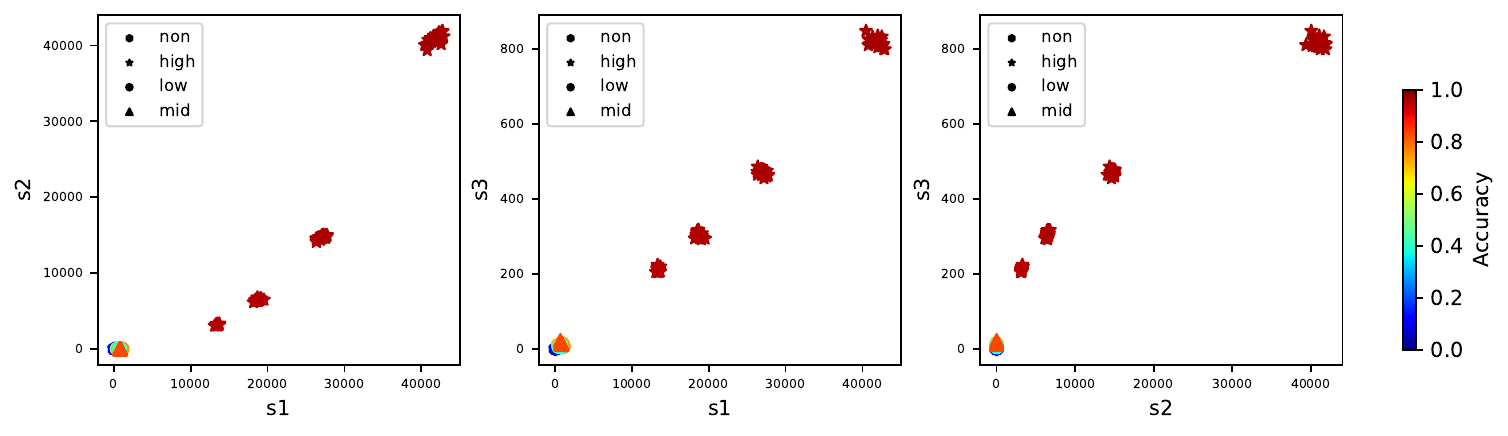}} \\
        \rotatebox[origin=c]{90}{CIFAR-10} & \raisebox{-0.5\height}{\includegraphics[width=0.80\textwidth]{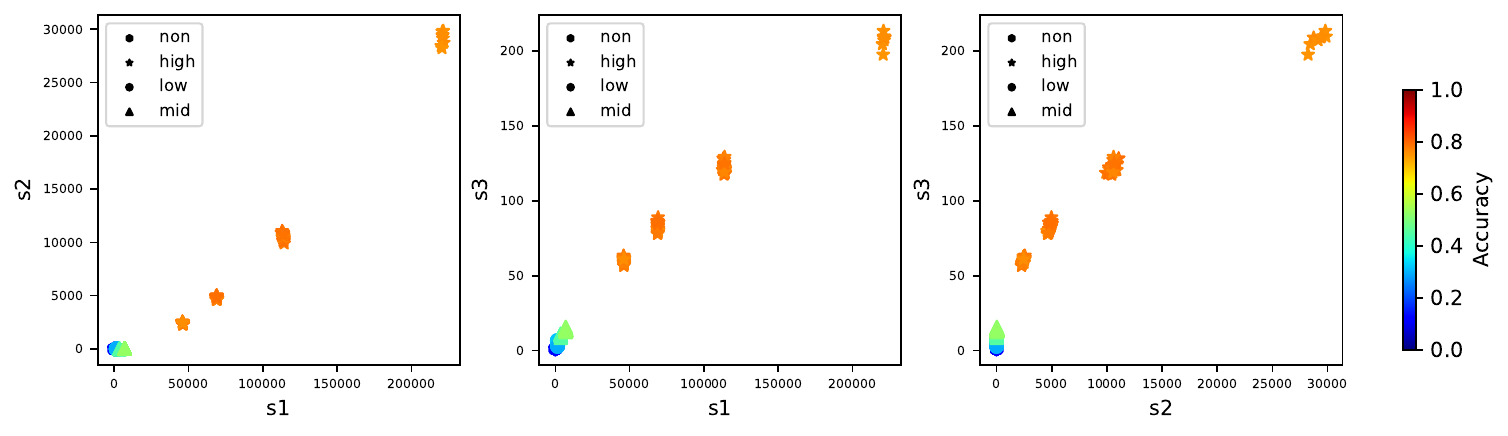}} \\
    \end{tabular}
    \caption{DNN node strength based characterization of optimal and suboptimal networks.  
    }
    \label{fig:dnn_weight_strength}
\end{figure}

\begin{figure}
    \centering
    \fontsize{7pt}{7pt}\selectfont
    \setlength{\tabcolsep}{5pt}
    \centering 
    \begin{tabular}{ll}
        & \begin{tabular}{cccc}
            \qquad Conv1 vs FC & \qquad\quad Conv1$^{+}$ vs FC$^{+}$ & \qquad\quad Conv1$^{-}$ vs FC$^{-}$ \\
        \end{tabular} \\
        \rotatebox[origin=c]{90}{MNIST} & \raisebox{-0.5\height}{\includegraphics[width=0.80\textwidth]{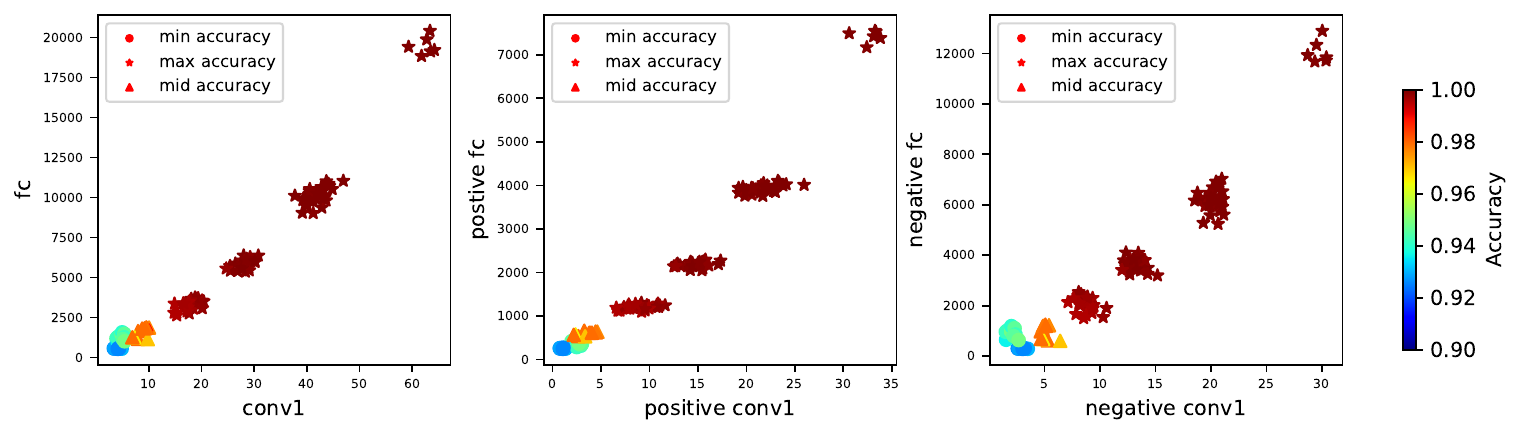}} \\
        \rotatebox[origin=c]{90}{FMNIST} & \raisebox{-0.5\height}{\includegraphics[width=0.80\textwidth]{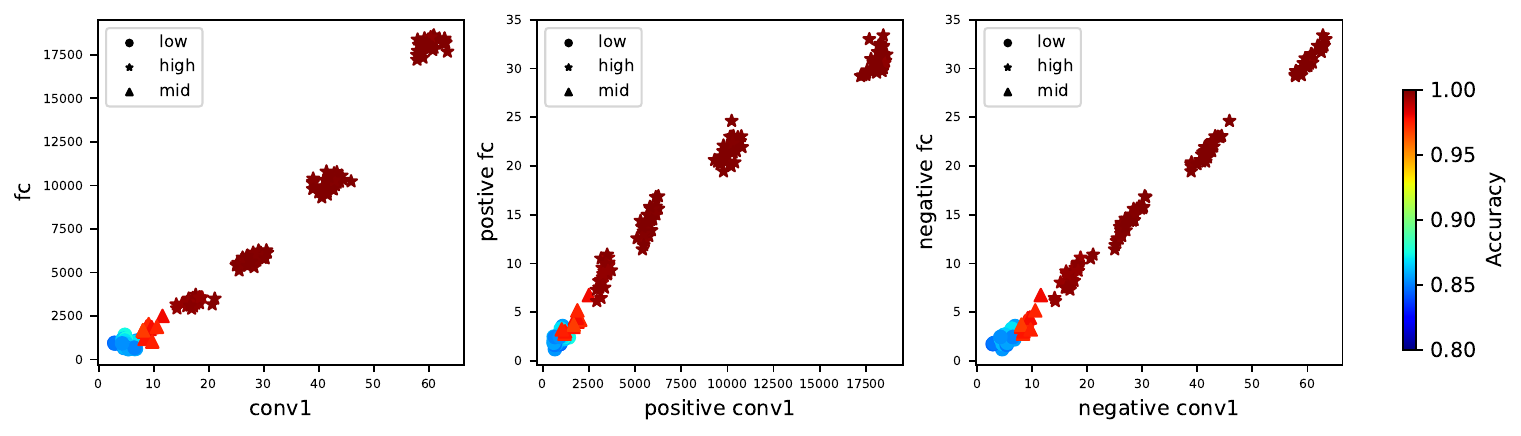}} \\
        \rotatebox[origin=c]{90}{CIFAR-10} & \raisebox{-0.5\height}{\includegraphics[width=0.80\textwidth]{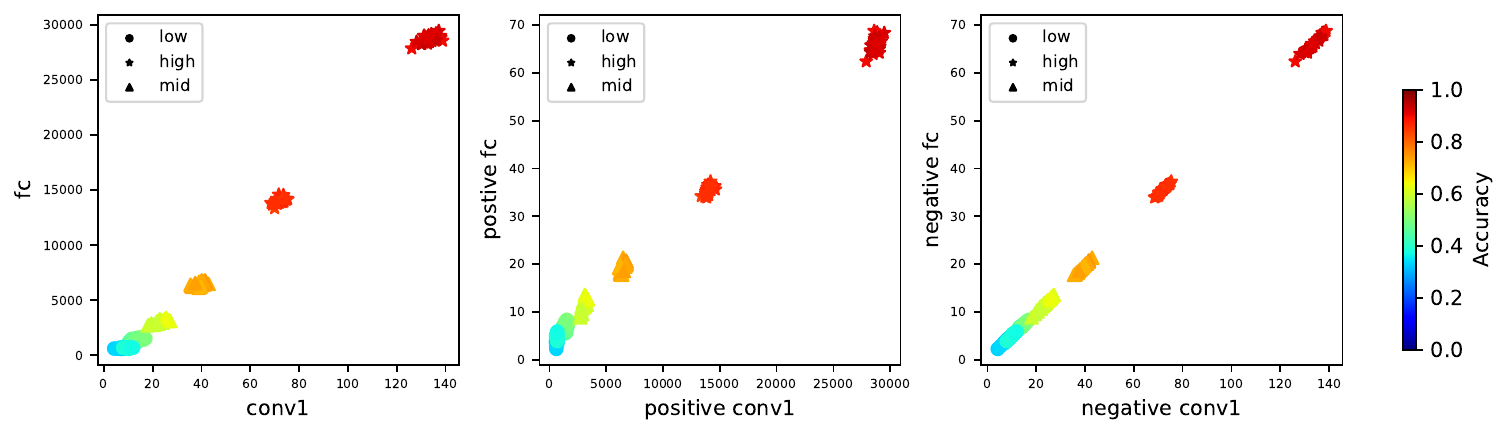}} \\
    \end{tabular}
    \caption{CNN node strength based characterization of optimal and suboptimal networks.}
    \label{fig:cnn_weight_strength}
\end{figure}

\begin{figure}
    \centering
    \fontsize{7pt}{7pt}\selectfont
    \setlength{\tabcolsep}{5pt}
    \centering 
    \begin{tabular}{ll}
        & \begin{tabular}{cccc}
            \qquad Attn vs MLP & \qquad MLP vs Norm & \qquad Attn vs Norm \\
        \end{tabular} \\
        \rotatebox[origin=c]{90}{MNIST} & \raisebox{-0.5\height}{\includegraphics[width=0.75\textwidth]{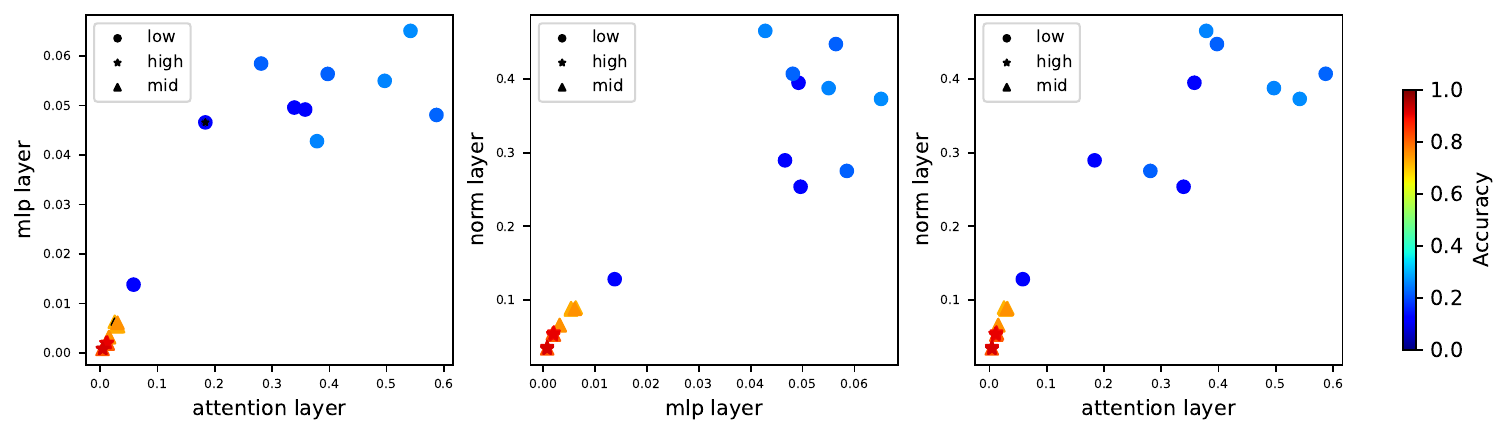}} \\
        \rotatebox[origin=c]{90}{FMNIST} & \raisebox{-0.5\height}{\includegraphics[width=0.75\textwidth]{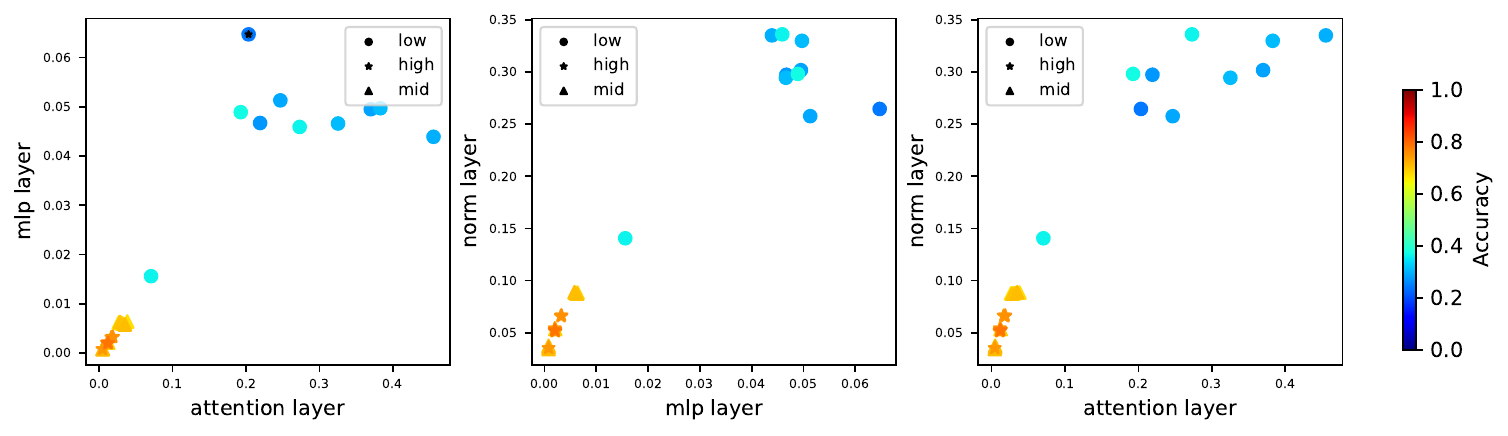}} \\
        \rotatebox[origin=c]{90}{CIFAR-10} & \raisebox{-0.5\height}{\includegraphics[width=0.75\textwidth]{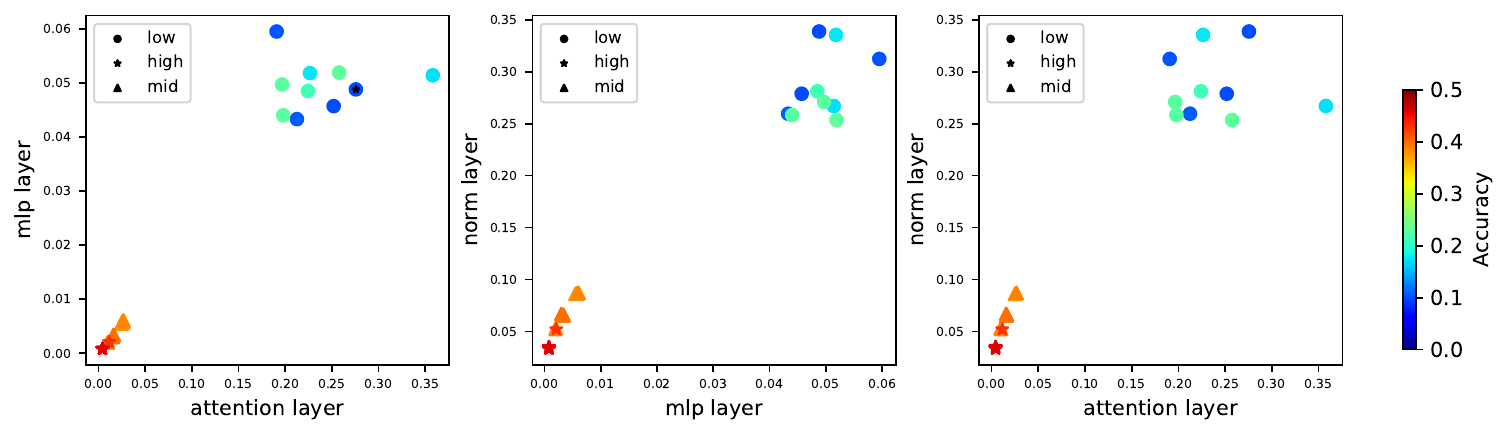}} \\
    \end{tabular}
    \caption{ViT node strength based characterization of optimal and suboptimal networks.}
    \label{fig:tf_weight_strength}
\end{figure}

\begin{figure}
    \centering
    \fontsize{7pt}{7pt}\selectfont
    \setlength{\tabcolsep}{5pt}
    \centering 
    \begin{tabular}{ll}
        & \begin{tabular}{cccc}
            \qquad I/P - FC1 & \qquad\qquad  FC1 - FC2 & \qquad\qquad FC2 - O/P \\
        \end{tabular} \\
        \rotatebox[origin=c]{90}{MNIST} & \raisebox{-0.5\height}{\includegraphics[width=0.75\textwidth]{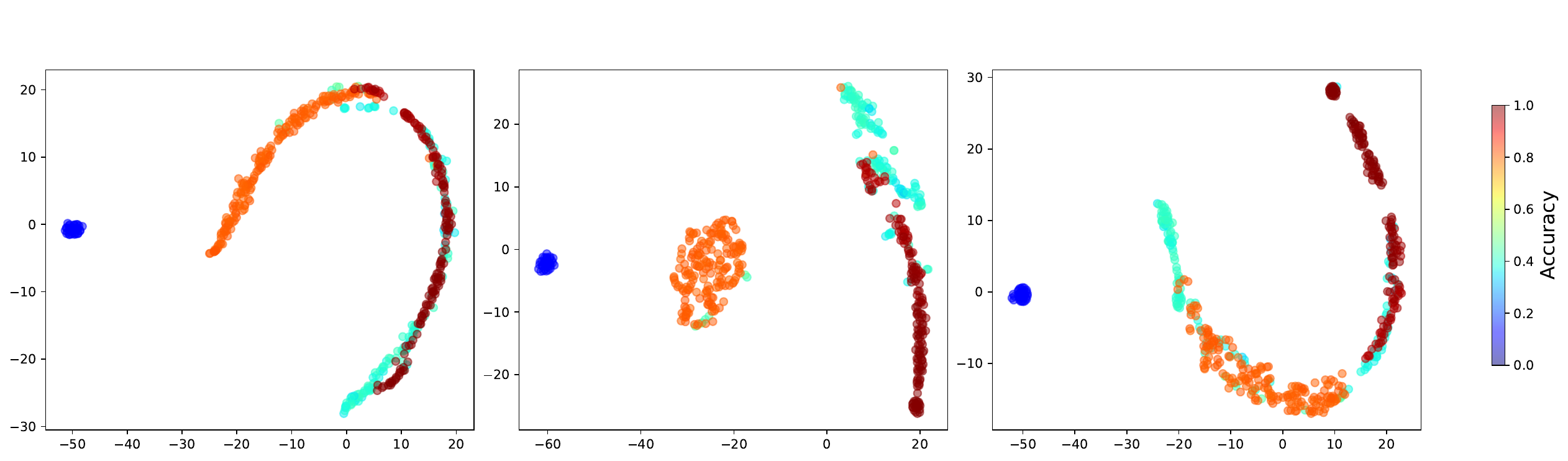}} \\
        \rotatebox[origin=c]{90}{FMNIST} & \raisebox{-0.5\height}{\includegraphics[width=0.75\textwidth]{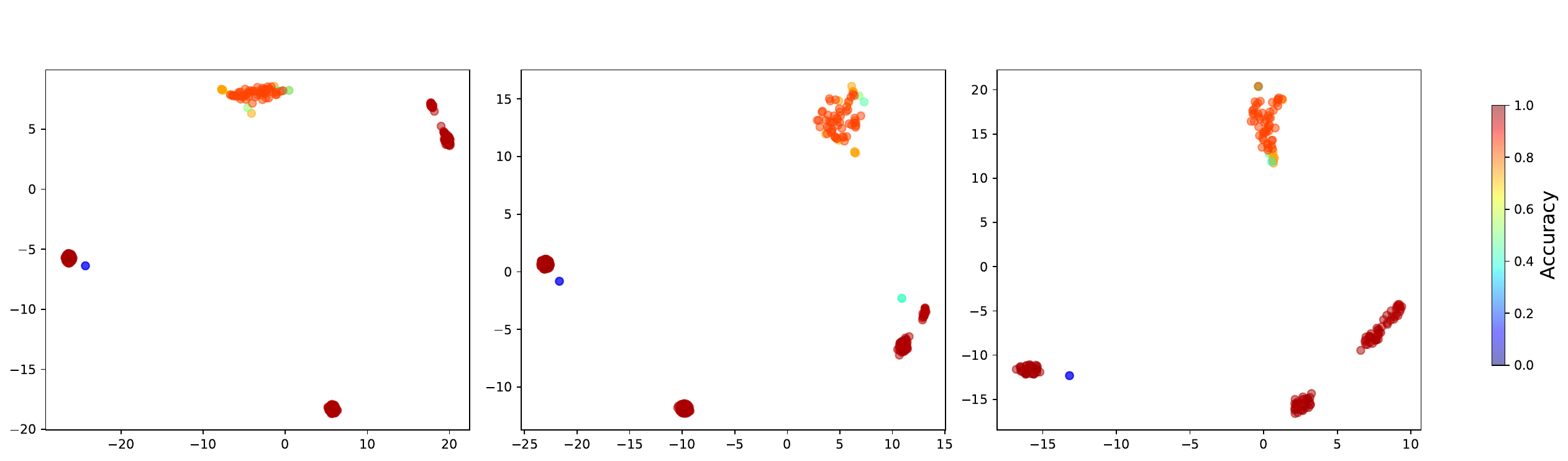}} \\
        \rotatebox[origin=c]{90}{CIFAR-10} & \raisebox{-0.5\height}{\includegraphics[width=0.75\textwidth]{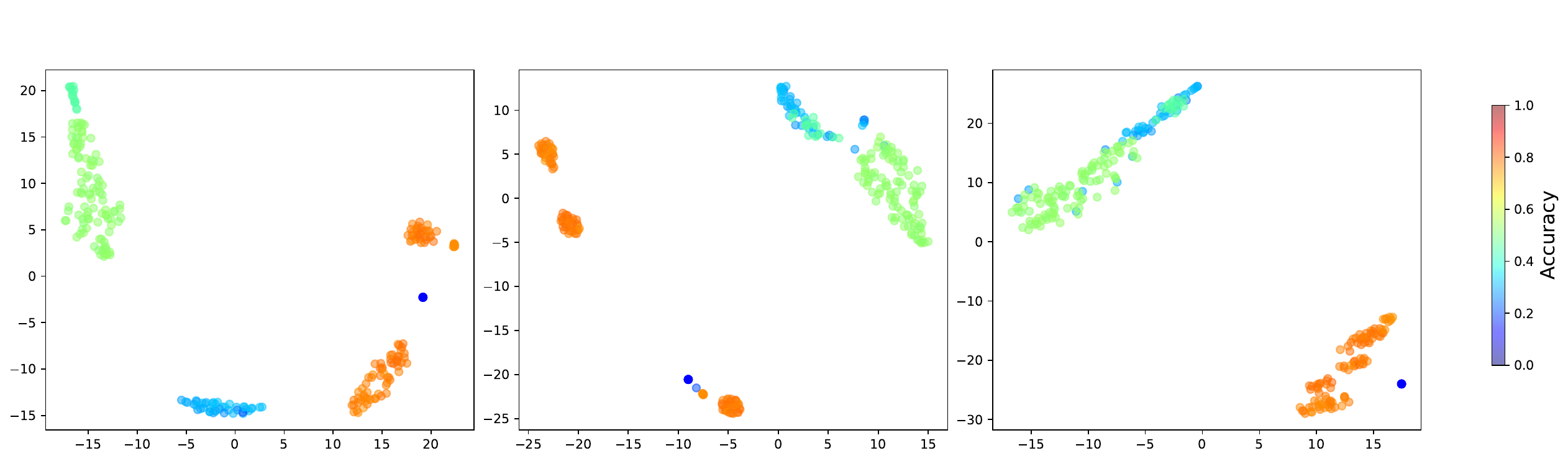}} \\
    \end{tabular}
    \caption{DNN network weight projection. 
    Optimal and suboptimal learnable parameter clusters are distinctly shown in accuracy colors. Less complex data has a smoother transition, and higher complex data has a sharp separation between clusters.
    }
    \label{fig:dnn_TSNE}
\end{figure}

\begin{figure}
    \centering
    \fontsize{7pt}{7pt}\selectfont
    \setlength{\tabcolsep}{5pt}
    \centering 
    \begin{tabular}{ll}
        & \begin{tabular}{cc}  
            ~\qquad ~~Conv1 & ~\qquad~\qquad\qquad ~~FC
        \end{tabular} \\
        \rotatebox[origin=c]{90}{MNIST} & 
           \begin{tabular}{cc} 
             \includegraphics[width=0.55\textwidth]{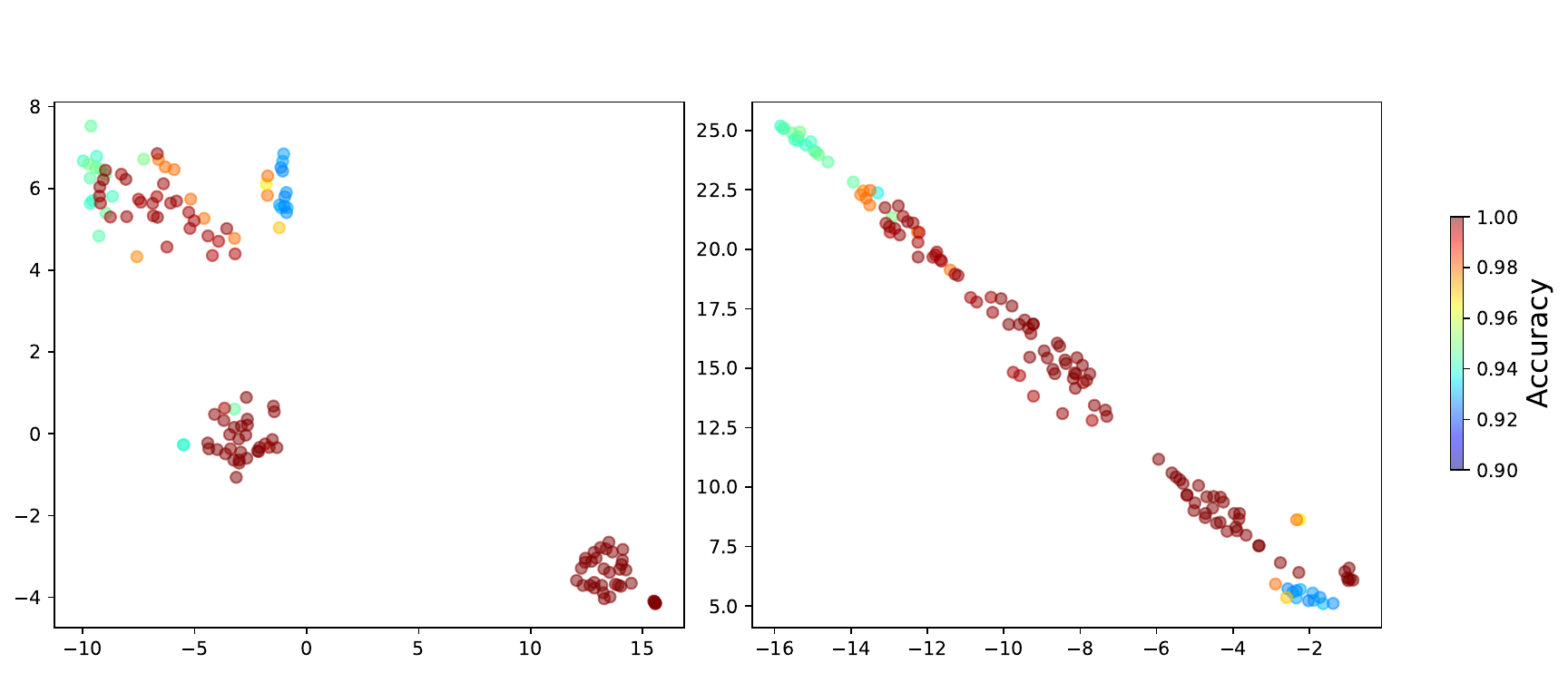} & \includegraphics[width=0.16\textwidth]{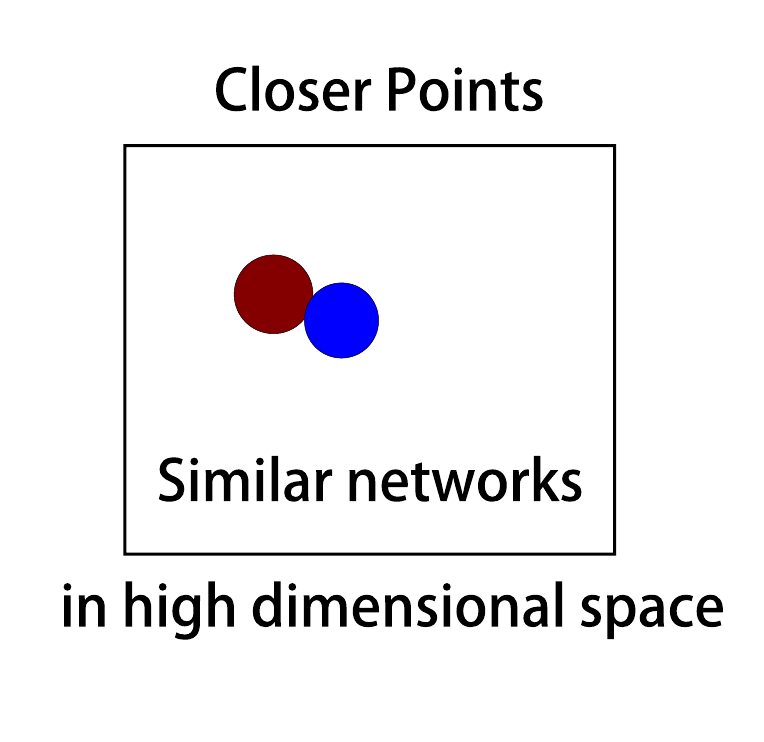}
           \end{tabular} \\
        \rotatebox[origin=c]{90}{FMNIST} & 
           \begin{tabular}{cc} 
             \includegraphics[width=0.55\textwidth]{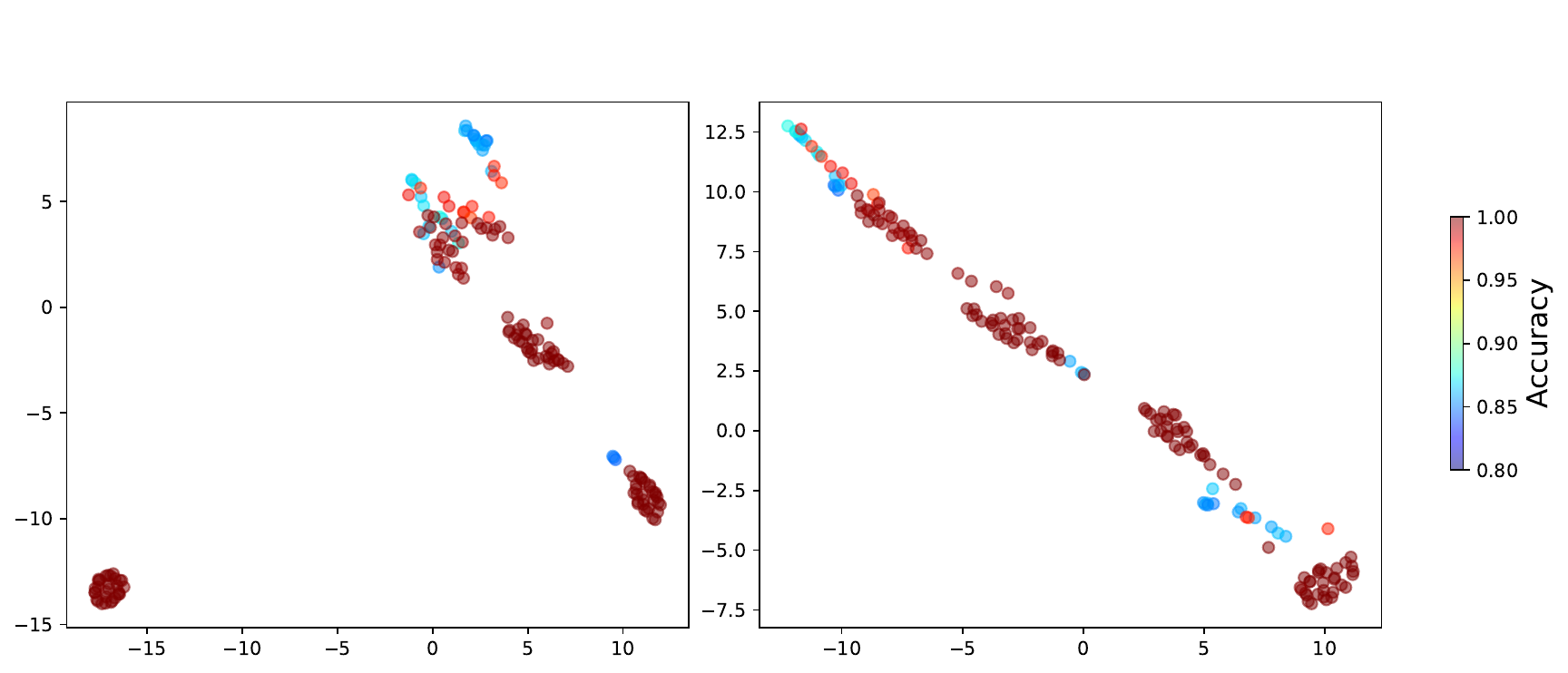} & \includegraphics[width=0.16\textwidth]{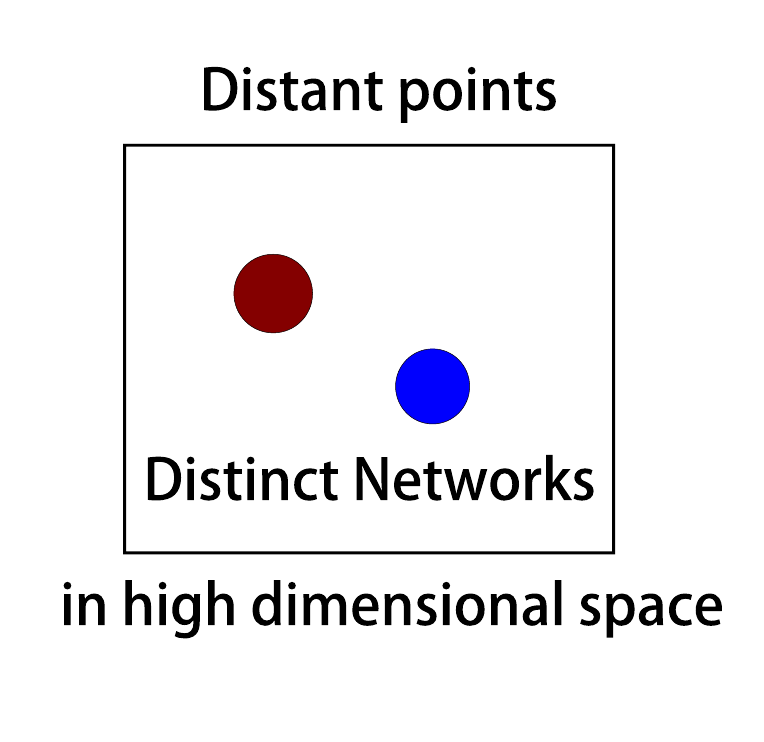}
           \end{tabular} \\
        \rotatebox[origin=c]{90}{CIFAR-10} & 
           \begin{tabular}{cc} 
             \includegraphics[width=0.55\textwidth]{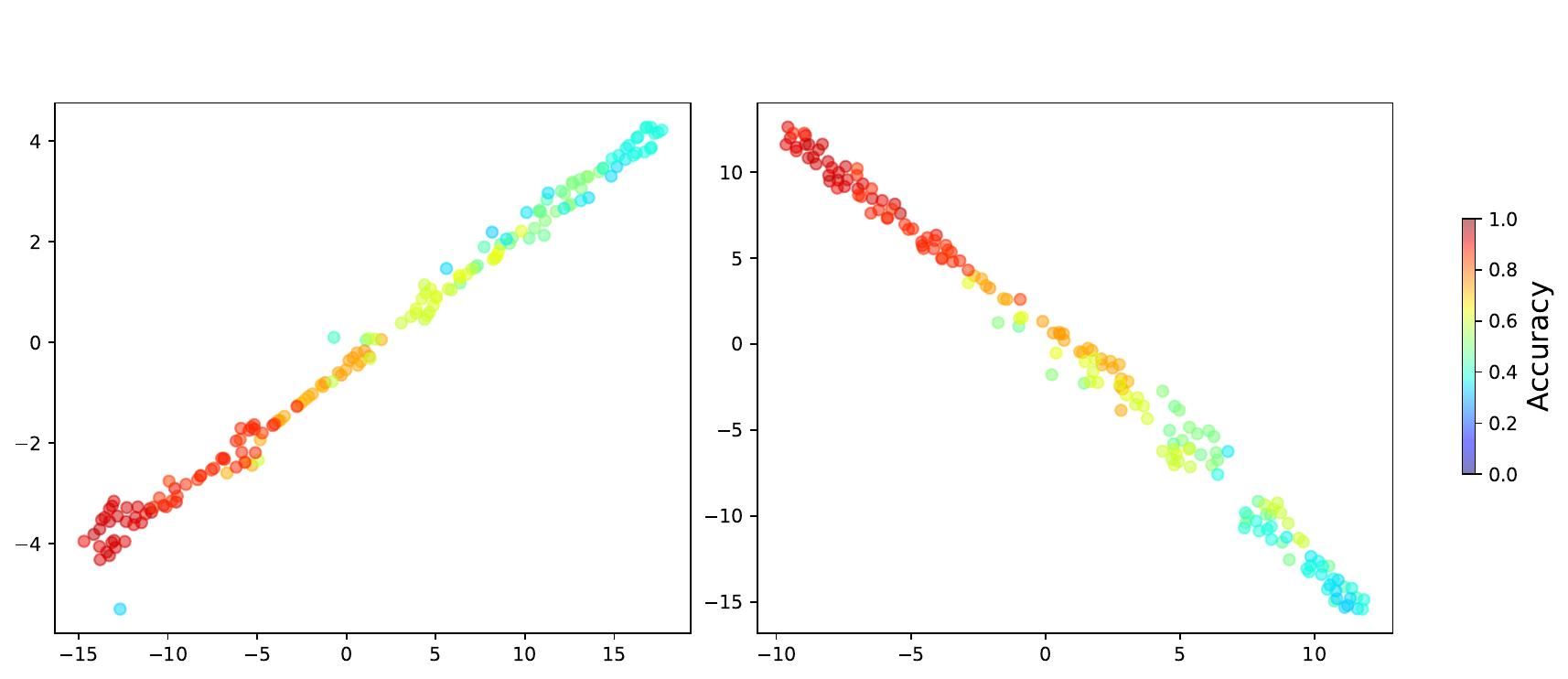} & \includegraphics[width=0.15\textwidth]{fig/colorbar.pdf}
           \end{tabular} \\
    \end{tabular}
    \caption{CNN network weight projection. 
    }
    \label{fig:cnn_TSNE}
\end{figure}

\begin{figure}
    \centering
    \fontsize{7pt}{7pt}\selectfont
    \setlength{\tabcolsep}{5pt}
    \centering 
    \begin{tabular}{ll}
        & \begin{tabular}{ccc}
            \qquad\quad Attn & \qquad\qquad\qquad\quad MLP & \qquad\qquad\qquad Norm 
        \end{tabular} \\
        \rotatebox[origin=c]{90}{MNIST} & \raisebox{-0.5\height}{\includegraphics[width=0.75\textwidth]{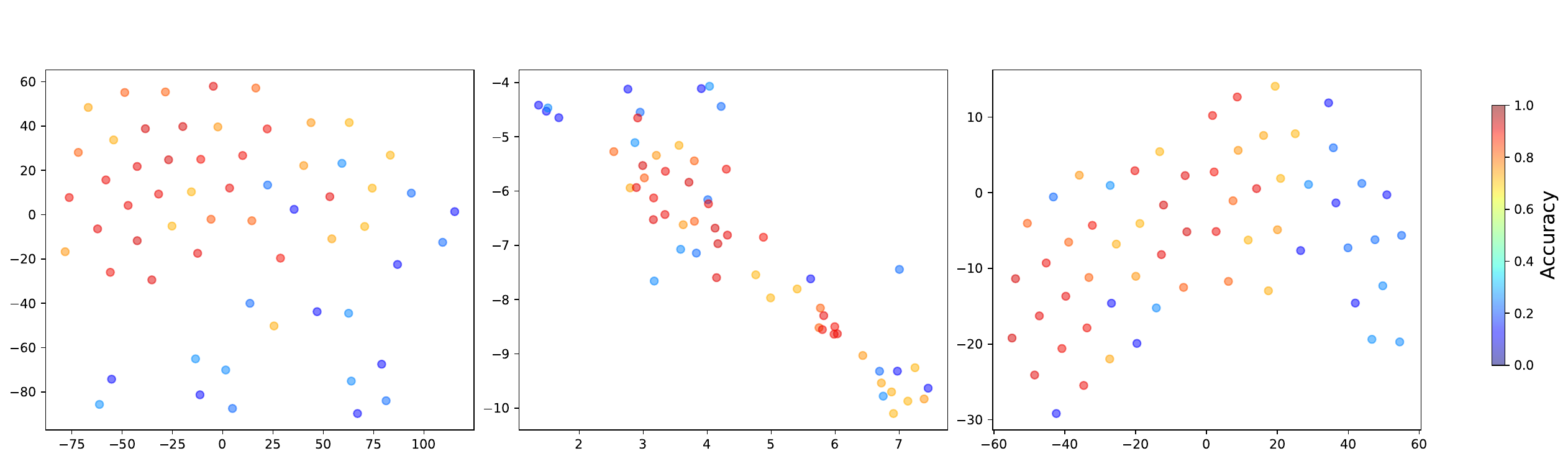}} \\
        \rotatebox[origin=c]{90}{FMNIST} & \raisebox{-0.5\height}{\includegraphics[width=0.75\textwidth]{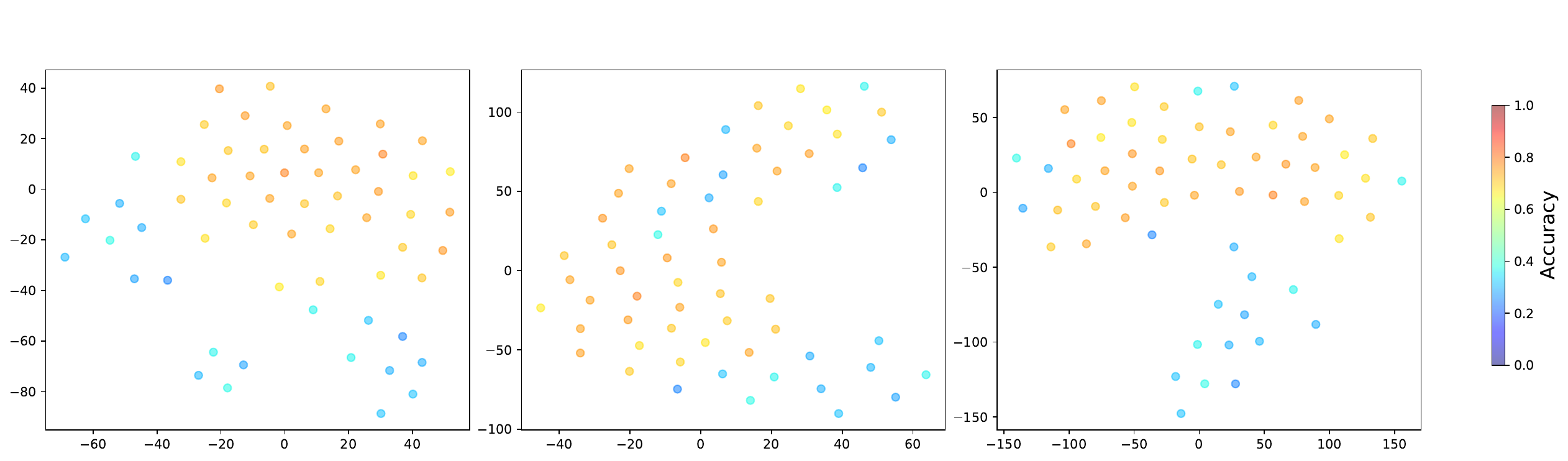}} \\
        \rotatebox[origin=c]{90}{CIFAR-10} & \raisebox{-0.5\height}{\includegraphics[width=0.75\textwidth]{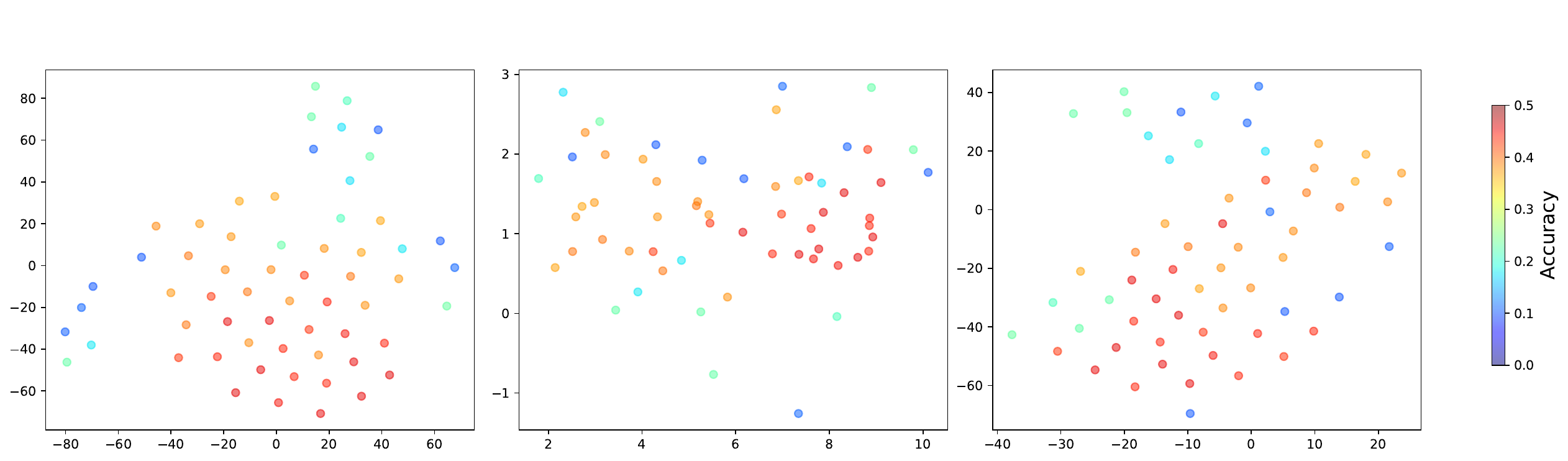}} \\
    \end{tabular}
    \caption{ViT network weight projection. 
    }
    \label{fig:tf_TSNE}
\end{figure}

We, therefore, present the characterization of the networks using the weight strength statistic analysis (in Figs.~\ref{fig:dnn_weight_meanstd}, \ref{fig:cnn_weight_meanstd}, and \ref{fig:tf_weight_meanstd}), weight distribution analysis using kernel density estimate (in Figs.~\ref{fig:dnn_weight_distribution}, \ref{fig:cnn_weight_distribution}, and \ref{fig:tf_weight_distribution}), node strength analysis (in Figs.~\ref{fig:dnn_weight_strength}, \ref{fig:cnn_weight_strength}, and \ref{fig:tf_weight_strength}), and weight projection analysis (in Figs.~\ref{fig:dnn_TSNE}, \ref{fig:cnn_TSNE}, and \ref{fig:tf_TSNE}): 

\textbf{Weight strength analysis}. Our comprehensive investigation using weight statistics (the mean (x-axis) and standard deviation (y-axis) of weight) in Figs.~\ref{fig:dnn_weight_meanstd}, \ref{fig:cnn_weight_meanstd}, and \ref{fig:tf_weight_meanstd} presents a compelling picture of the discrimination between optimal and suboptimal networks. High-accuracy networks (shown in dark red) in DNN, CNN, and ViT architectures show tight clustering of weights with low standard deviation, indicating a stable and efficient learning state, respectively. On the contrary, we observe that the low-accuracy and mid-accuracy networks show high variance in their performances, indicating unstable weight convergence.

Specifically, we observe that this variance in the successful network's weight decreases in the final/classification layer (clearly observed in DNNs and CNNs final FC layer), indicating that the high-accuracy network has finely optimized weight belonging to a particular distribution, which is crucial to enhancing correct feature detection and classification. In contrast, when networks fail, their weight distribution is scattered, indicating the networks are prematurely stuck at suboptimal/local optimal space. In this context, we also analyzed weight distribution patterns across different accuracy groups in Figs.~\ref{fig:dnn_weight_distribution}, \ref{fig:cnn_weight_distribution}, and \ref{fig:tf_weight_distribution}, revealing a clear peak near zero for high-accuracy models, indicative of their ability to emphasize crucial features while minimizing noise. This contrasts with the flatter distributions seen in mid and low-accuracy models, which suggests less efficient feature discrimination and potential overfitting to non-essential features.

\textbf{Node strength analysis}. We investigate the node strengths between layers in Figs.~\ref{fig:dnn_weight_strength}, \ref{fig:cnn_weight_strength}, and \ref{fig:tf_weight_strength} to assess whether the successful networks show similar performance as weight strength analysis. We observe clear correlations in high-accuracy networks, indicating robust inter-layer communication and efficient signal propagation. This is especially evident in DNN and CNN networks, where the node strength of layers correlates strongly. In ViT, the strength values cluster and correlate with the high-accuracy networks. However, contrary to the DNN and CNN networks, the strength values for the high-accuracy network are low for high-accuracy networks, inciting the lower concentrated values in the ViT attention layer, and the MLP layer in ViT is better for high-performing.

\textbf{Weight projection analysis}. We investigate the weight projection in Figs.~\ref{fig:dnn_TSNE}, \ref{fig:cnn_TSNE}, and \ref{fig:tf_TSNE}, which is equivalent to projecting high-dimensional network weight vectors to a two-dimensional space using t-SNE. This projection shows the position/proximity of different networks on a high-dimensional space~\cite{rauber2016visualizing}. We assess whether trained networks cluster together or not. We observe that high-accuracy networks have their layers weights clustered distinctly separately compared to the low/mid-accuracy networks. For ViT, due to the low number of runs/experiments compared to DNN and CNN, the projection is sparse. 

\section{Conclusion}\label{sec:conlusion}
We present a methodology to characterize deep learning uncertainty of success and failure by comprehensively analyzing learnable parameters (weights) strength, node strength, and weight projection of networks on three models: deep neural networks, convolutional neural networks, and vision transformers over three datasets: MNIST, Fashion MNIST, and CIFAR-10. Our finding reveals that successful networks have low variance in their weight, and their weight converges to a similar weight distribution, clustering close to each other in high-dimensional space. On the other hand, failed networks show contrary behavior to successful networks, i.e., they have large variances in their weights and appear to be further away from the successful cluster. The node strength of a successful network has a tendency to increase strength values for DNNs and CNNs.

\bibliographystyle{splncs04}
\bibliography{icannref} 

\end{document}